\documentclass{article}

\usepackage{neurips}
\usepackage[utf8]{inputenc} 
\usepackage[T1]{fontenc}    
\usepackage{hyperref}       
\usepackage{url}            
\usepackage{booktabs}       
\usepackage{amsfonts}       
\usepackage{nicefrac}       
\usepackage{microtype}      
\usepackage{xcolor}         
\usepackage{amsmath}
\usepackage{subcaption}
\usepackage{graphicx}

\title{LatentPINNs: Generative physics-informed neural networks via a latent representation learning}

\author{
Mohammad H. Taufik \\
Department of Physical Science and Engineering\\
King Abdullah University of Science and Technology\\
Thuwal, 23955, Saudi Arabia\\
\texttt{mohammad.taufik@kaust.edu.sa.}\\
\And
Tariq Alkhalifah\\
Department of Physical Science and Engineering\\
King Abdullah University of Science and Technology\\
Thuwal 23955, Saudi Arabia\\
\texttt{tariq.alkhalifah@kaust.edu.sa}\\
}

\begin{document}

\maketitle
\begin{abstract}
Physics-informed neural networks (PINNs) are promising to replace conventional partial differential equation (PDE) solvers by offering more accurate and flexible PDE solutions. However, they are hampered by the relatively slow convergence and the need to perform additional, potentially expensive, training for different PDE parameters. To solve this limitation, we introduce latentPINN, a framework that utilizes latent representations of the PDE parameters as additional (to the coordinates) inputs into PINNs and allows for training over the distribution of these parameters. Motivated by the recent progress on generative models, we promote the use of latent diffusion models to learn compressed latent representations of the PDE parameters distribution and act as input parameters to NN functional solutions. We use a two-stage training scheme in which the first stage, we learn the latent representations for the distribution of PDE parameters. In the second stage, we train a physics-informed neural network over inputs given by randomly drawn samples from the coordinate space within the solution domain and samples from the learned latent representation of the PDE parameters. We test the approach on a class of level set equations given by the nonlinear Eikonal equation. We specifically share results corresponding to three different sets of Eikonal parameters (velocity models). The proposed method performs well on new phase velocity models without the need for any additional training.
\end{abstract}
\section{Introduction}

Partial differential equations (PDEs) provide mathematical formulations that describe many physical systems. They have been the cornerstone of scientific and engineering applications for decades. In practice, these equations are often discretized and used to solve for the state field or the unknown physical parameters (e.g., velocity, pressure, heat, etc.). With complex boundaries and large computational domains, conventional PDE solvers developed over the years (e.g., finite-difference and finite element methods) suffer from general inefficiency and require complex discretization procedures. Neural network-based PDE solvers have shown considerable promise in addressing these limitations, as well as providing more efficient solutions.  In this family of approaches, neural networks can be utilized to either learn the governing PDE operator (neural operators) \cite[]{li2020fourier,lu2019deeponet,li2020neural,kovachki2021neural} or be regularized (by the governing PDE) to learn a specific PDE solution (neural instance solvers) \cite[]{raissi2019physics,sirignano2018dgm}. Neural operators often require the solution labels (obtained, for example, from conventional numerical solvers), in which the training process is then performed in a supervised manner. Thus, at inference time, neural operators can accommodate different realizations of the PDEs. This comes, however, with heavy computational costs for large-scale problems, as in this case, acquiring the labels themselves might require significant computing power. Several attempts have been made to relax the data-driven nature of neural operators by incorporating physical laws into the neural operators \cite[]{li2021physics}; these additional constraints, however, still render the solution mesh dependent and generally inefficient to train. One notable example of neural instance solvers, termed physics-informed neural networks (PINNs), on the other hand, is driven by the governing PDE at hand (independent of numerical PDE solvers). At inference time, it provides the functional representation for a specific PDE parameter. These parametric deep neural networks (DNNs) address the limitations of numerical methods by facilitating flexible (mesh-independent) solutions that are generally more accurate than their numerical counterparts, thanks to the functional nature of the derivative evaluations.

In this quest to fully replace conventional PDE solvers, the current PINNs are facing two main challenges. Firstly, randomly initialized PINNs often require thousands of iterations (epochs) to converge, massively increasing the cost of attaining the PDE solution. To this end, several attempts have been made to address the training dynamics as well as improve its efficiency. \cite{fang2021high} promoted the use of a local fitting approach to approximate the governing PDE operator and improve PINNs' efficiency. \cite{wang2022respecting} proposed the use of causality constraint that results in better convergence of PINNs from some initial condition. \cite{huang2022pinnup} promoted the use of higher-dimensional embedding space of the coordinate input to improve PINNs' convergence. \cite{huang2023efficient,rzepecki2022fast} used alternative embedding layers and, thus, demonstrated significant computational speed-ups. Moreover, a slow convergence can also be attributed to the challenging training dynamics. One major challenge comes from the need to appropriately balance the various loss terms during the PINNs training. \cite{schiassi_extreme_2021,huang2023microseismic,taufik2023stable} promote the use of hard-constrained data loss function to alleviate the burden of this balancing procedure. Secondly, PINNs are trained for specific PDE parameters; a different parameter set requires additional PINN training. To accommodate for different PDE parameters efficiently, several authors suggest the use of transfer learning \cite[]{goswami2020transfer,bin2021pinneik} and meta learning \cite{penwarden2021physics,liu2022novel,huang2022meta}. For transfer learning to reduce the cost of training effectively, the difference between the new PDE parameters and the original ones in which the PINN was trained needs to be small. Such a restrictive condition might end up requiring almost identical training costs to that from randomly initialized PINNs. Meta learning, on the other hand, comes with the objective of finding the best starting point to accelerate the new PDE parameter PINNs training. Both of these approaches, unlike neural operators, still require additional training for new PDE parameters.

In this paper, we introduce the use of latent representation learning to PINNs. Informed with this additional latent representation, the trained PINNs can be regarded as generative models that produce solutions to PDEs as a function of latent variables (representing the PDE parameters). We investigate the possibility of overcoming additional training for new PDE parameters using a pre-training and fine-tuning scheme. In the experiments, we test our framework on a nonlinear PDE with various datasets and share the challenges we faced.

The main contributions of this paper can be summarized as follows:

\begin{itemize}
    \item We introduce the use of latent representation learning to physics-informed neural networks.
    \item We promote the use of autoencoder-based models to learn the latent representation of PDE parameters that can be combined with state-of-the-art diffusion models for functional data representation.
    \item We test the proposed algorithm on three datasets to solve a nonlinear PDE without the need to perform additional training for new PDE parameters.
\end{itemize}

The organization of the paper is as follows. We first discuss the related works in the next section. We then provide the theoretical background and the proposed framework in Section 3, followed by numerical experiments in Section 4. Finally, we discuss the current challenges and future directions, as well as conclude in the last two sections.
\section{Related Work}


\subsection{Generalizing PINNs}

There have been many attempts to pre-train PINNs so that they converge faster for new PDE parameters. Transfer learning was suggested in which the PINNs are initialized with the weights of the training for a previous set of PDE parameters \cite[]{bin2021pinneik}. In transfer learning, the change in the PDE parameters needs to be small to improve convergence compared to starting with random initialization. Alternatively, meta learning was adopted in PINN training by finding initial weights that can work for a range of tasks or PDE parameters \cite{penwarden2021physics,liu2022novel,huang2022meta}. Meta learning is, however, costly considering the pretraining on a range of parameters. In all of these methods, additional training is required to fine-tune the PINN to provide accurate solutions for a new set of PDE parameters. In other words, these methods do not necessarily reduce the requirement for an extra PINNs training step for different PDE parameters.

\subsection{Generative models with PINNs}

Uncertainty estimation has been the primary goal of incorporating generative models into physics-informed neural networks \cite[]{yang2018physics,zhu2019physics,yang2020BPINNsBP,linka2022BayesianPN,oszkinat2022uncertainty}. For such purposes, the generative models are utilized such that they can stochasticity be included in the \emph{deterministic} vanilla PINNs. \cite{huang2022meta} introduces the use of an auto-decoder network to accelerate PINNs training from a meta-learning perspective. Different PDE parameters can be viewed as different tasks. Though latent representation is also included as an additional input parameter, the trained PINN still requires additional training for new PDE parameters. From the perspective of reduced order modeling, \cite{kim2022fast} promotes the use of a masked autoencoder to learn the lower-dimensional manifold of the PDE parameters. They use the trained decoder and incorporate existing conventional PDE solvers instead of neural networks to solve the PDE. Finally, \cite{zou2023hydra} introduces the use of normalizing flow to perform density estimation for further downstream tasks. Extra training for new PDE parameters, however, is still required for those tasks.

\subsection{Incorporating latent vectors into PDE solvers}

Almost any machine learning algorithm inherently tries to learn the hidden \emph{latent} representation between the input data that can be mapped into an output for a certain task. In a more restrictive manner, these latent variables can be constructed (and later accessed) such that they represent the data with a significant reduction in size. \cite{ranade2021latent} incorporates latent vectors with existing numerical solvers to generalize the PINNs solutions. The idea of operating on a much lower-dimensional space for solving PDEs has been around under the banner of model order reduction \cite[]{lumley1967structure}. Motivated by this idea, \cite{wu2022learning} accelerates PINNs for solving large-scale time-dependent problems. Similarly, \cite{kontolati2023learning} incorporates such an idea into a class of neural operators to achieve better accuracy.

To the best of our knowledge, we promote the first attempt to make PINNs learn solutions for a certain PDE parameter distribution without the need for additional training to accommodate new PDE parameters within the distribution. In such an approach, we aim to incorporate a generative feature into the trained PINNs model, such that it can generate various PDE solutions for various PDE parameters.
\section{Methodology}

Consider a general parametric PDE in the form of

\begin{equation}
u_t(\boldsymbol{x}, t ; \boldsymbol{\xi})+\mathcal{A}(u(\boldsymbol{x}, t ; \boldsymbol{\xi}))=\mathcal{S}(\boldsymbol{x}, t ; \boldsymbol{\xi}), \quad \mathcal{B}(u ; \boldsymbol{\xi})=0, \quad t \in [0, T], \quad \boldsymbol{x} \in \Omega, \quad \boldsymbol{\xi} \in \mathcal{P},
\end{equation}

where $u(\boldsymbol{x}, t; \boldsymbol{\xi})$ denotes the desired field as a function of the spatio-temporal variables $[\boldsymbol{x}, t]$, in which $u_t$ corresponds to its first-order temporal derivative. Here, $\boldsymbol{\xi}$ denotes the PDE parameters, and $\Omega$ is a subset of a $d$-dimensional domain of real numbers, $\mathbb{R}^d$. $\mathcal{A}$ and $\mathcal{F}$ denote the non-linear operator and the source function, respectively, applied on the computational domain $\Omega$, while $\mathcal{B}$ defines the boundary operator applied on $\partial \Omega$ (the surrounding surface). Conventional PINNs will then try to learn a mapping between the input ($\boldsymbol{x}, t$) to the PDE solution $u$ for that input location. This mapping is facilitated by a neural network $\mathcal{N}(\boldsymbol{x}, t ; \boldsymbol{\xi}, \boldsymbol{\theta})$, where $\boldsymbol{\theta}$ denotes the network's parameters. The output of this network will then be regarded as a surrogate to the PDE solution $\hat{u}$. During the PINNs' training, the PDE residuals and their boundary conditions are incorporated into the objective function in the form of

\begin{align}
\mathcal{L} &= \mathcal{L}_{\text{PDE}} + \mathcal{L}_{\text{boundary}}, \\
&= ||\hat{u}_t(\boldsymbol{x}, t ; \boldsymbol{\xi})+\mathcal{A}(\hat{u}(\boldsymbol{x}, t ; \boldsymbol{\xi}))-\mathcal{S}(\boldsymbol{x}, t ; \boldsymbol{\xi})||_2^2 + ||\mathcal{B}(\hat{u}(\boldsymbol{x}, t ; \boldsymbol{\xi}))||_2^2.    
\end{align}

To this end, once the training is performed for specific PDE parameters $\boldsymbol{\xi}$, additional training is required (either by transfer learning or within a meta learning framework) to obtain PDE solutions for other PDE parameters.

\subsection{Latent Representation Learning}

The objective of most representation learning algorithms is to find a compressed representation of a dataset without losing any of its key information that distinguish the samples in the dataset. Given an input $\textbf{x}$, the objective function used for representation leanring can be broadly formulated as
\begin{align}
\mathcal{L} &= \mathcal{L}_{\text{reconstruction}} + \mathcal{L}_{\text{regularization}}.    
\end{align}
Regularization is often introduced to balance the trade-off between semantic and perceptual compression \cite[]{esser2021taming}. The reconstruction term can come in many flavors depending on the choice of the network architecture. In the simplest case, autoencoder networks \cite[]{rumelhart1985learning,hinton2006reducing} utilize deep neural networks to both encode (the input data) and decode (the latent vector). Though they perform well in dimensionality reduction, the learned latent vectors do not represent the data distribution well for generative tasks. To solve this limitation, variational autoencoders \cite[]{kingma2013auto} resort to a Bayesian representation of the latent vector. Instead of mapping between vectors, they map the input vectors to their latent distributions. This results in a better learned posterior (data) distribution. Generative adversarial networks (GANs) \cite[]{goodfellow2020generative} replace this encoder-decoder scheme with a discriminator-generator pair that are trained in an adversarial manner. This results in a more representative learned posterior distribution than the variational autoencoder at the cost of more challenging training dynamics (e.g., mode collapse pathology). The rise of score-based diffusion models \cite[]{song2019generative,song2020score,song2021maximum} in outperforming GANs in representing the data distribution inspired us also to utilize this model and incorporate it into physics-informed neural networks. So, we utilize such powerful generative models to introduce a generative feature into physics-informed neural networks.

\subsection{PINNs with latent space inputs}

Motivated by the recent success of latent-based models \cite[]{rombach2022high}, we introduce a two-stage training scheme into the PINNs training. During the first stage, a Kullback-Leibler divergence regularized autoencoder is utilized to learn the compressed latent representation of the PDE parameters. The resulting latent vectors (representing a Gaussian distribution mapping of the PDE parameters) will then be incorporated as an additional input into the PINNs training (second stage).

\begin{figure}[!ht]
    \centering
    \includegraphics[width=0.9\linewidth]{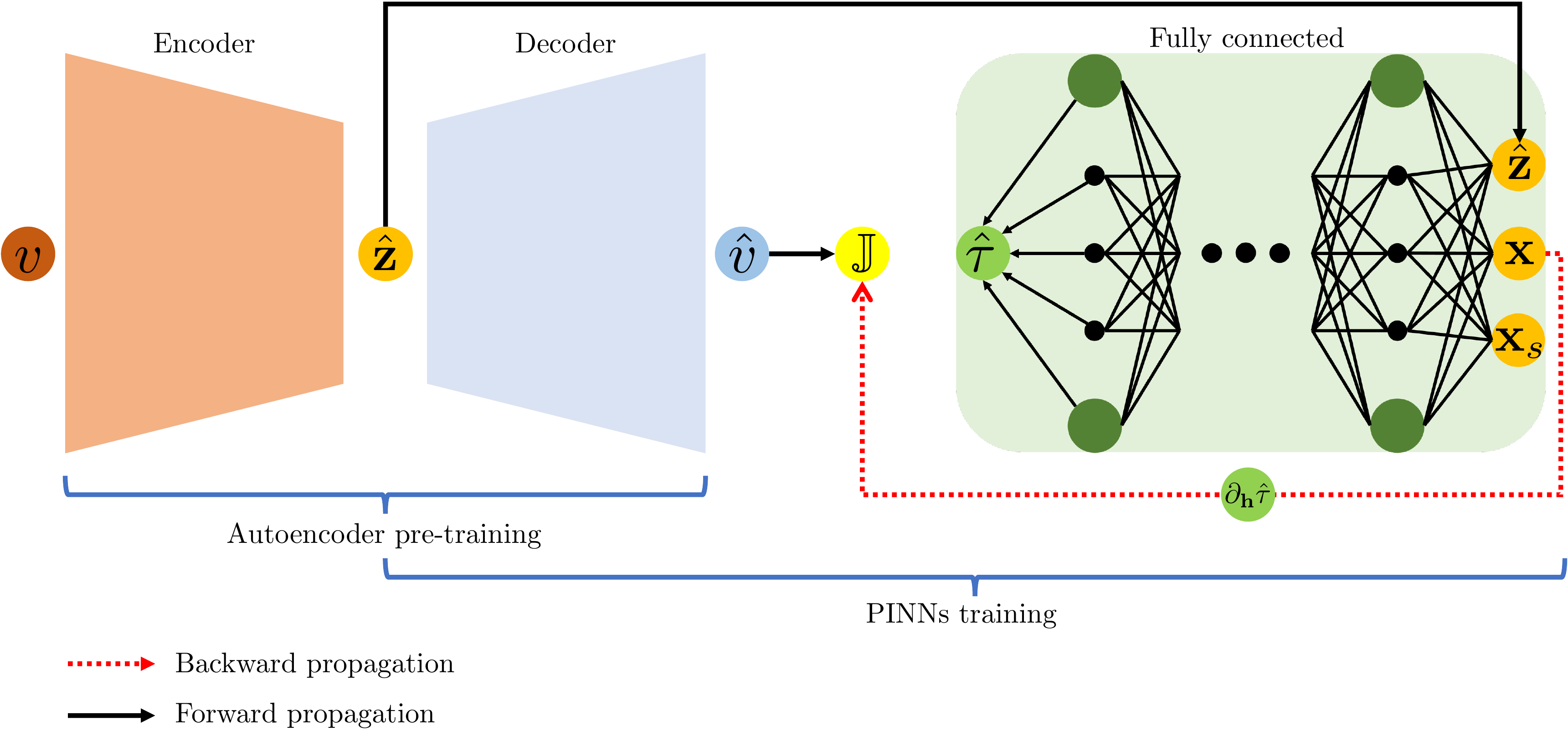}
    \caption{The proposed workflow of the latentPINN.}
    \label{workflow}
\end{figure}

Figure \ref{workflow} illustrates the overall workflow of the proposed method. As opposed to only using the position coordinates to train the PINNs \{$\textbf{x},\textbf{x}_s$\}, we also incorporate the latent vectors $\textbf{z}$ as an additional input parameter. Specifically, the multi-layer perceptron used as the main PINN model outputs the PDE solution denoted by $\hat{\tau}$. The PINN objective function, $\mathbb{J}$, is formulated by the PDE residual and parameterized by the lateral derivatives $\partial_h \hat{\tau}$ computed using backpropagation through the network. In this study, as we will see later, we consider modeling a solution for a class of level set equations given by the Eikonal equation. The objective then becomes predicting the PDE solution (traveltime fields) given knowledge of the PDE parameter (medium phase velocity fields $v$). We replace the true PDE parameters by the reconstruction result from the autoencoder $\hat{v}$.
\section{Numerical tests}

In this section, we describe the setup of the problem, specifically the partial differential equation we aim to solve and the neural network model used to solve the problem, including its training parameters. Finally, we share the results of testing our approach.

\subsection{The PDE problem}

We test the proposed method on a level-set-based PDE, often used
as a tool for numerical analysis of surfaces and shapes. Specifically, the level-set model often performs numerical computations for surfaces on a fixed Cartesian grid without the need to parameterize these objects \cite[]{OSHER198812}. Considering a surface, like a wavefront, moving with speed $V$ in the direction normal to the surface, the level set function $\phi$ satisfies the following nonlinear first-order PDE:
\begin{equation}  
\label{eqt0}
\frac{\partial \phi(\boldsymbol{x},T)}{\partial T} = V(\boldsymbol{x}) |\nabla \phi(\boldsymbol{x},T)| , \quad \boldsymbol{x} \in \mathbb{R}^3,
\end{equation}
where $T$ is the traveltime, and the operator $|.|$ is the Euclidean norm.

In wave theory, the level set function $\phi$ represents the phase of a wave, defining its geometrical shape in 2D or 3D as a function of time. In the high frequency asymptotic approximation, $\phi(\boldsymbol{x},T)= \omega T(\boldsymbol{x})$, and equation~\ref{eqt0} reduces
to its Eikonal form as follows
\begin{equation}  
\label{eqt1}
\left(\nabla T(\boldsymbol{x})\right)^2 = \frac{1}{V^2(\boldsymbol{x})}, \quad \boldsymbol{x} \in \mathbb{R}^3,
\end{equation}
where the solution of the PDE is represented by the traveltime field, $T$. Thus, the PDE is parameterized by the phase velocity of the medium $V$, which can be inhomogeneous.

In the following cases, we consider the problem of solving the Eikonal equation, which is equivalent to traveltime modeling for an inhomogeneous medium phase velocity model representing the PDE parameters. We, specifically, aim to solve the equation for a point source in a 2D Cartesian computational domain ($\textbf{x} \in \{x,z\} \in \mathbb{R}^2$). To mitigate source-related singularity, which is a known problem in solving eikonal equations numerically, we first substitute the traveltime field $T$ with $T_0\hat{\tau}$, known as the factorization approach. For details of its implementation in PINNs, we refer the readers to \cite{taufik2023neural}. Here, we consider a domain of size 5 km ($x \in [0,5]$) horizontally and 1 km ($z \in [0,1]$) vertically. The velocity models are discretized in this domain with 128 points along each axis. For all examples, the source is located in the middle ($\textbf{x}_s \in \{2.5,0.5\}$ km).

\subsection{The neural network setup for training}

As described earlier, we incorporate a two-stage training workflow for latentPINN before direct inference on new PDE parameters (velocity models). 
In the first stage, we train an autoencoder to learn a latent representation of the PDE parameters (velocity models), and in the second stage, we train a PINN that includes an added latent variables input. For the three examples below, we consider a latent dimension of 96 to be used on the bottleneck of the fully-connected layer.

The autoencoder used for the first stage of training is made up of an encoder part that takes three-channel images of size 128x128 pixels and performs downsampling across five convolutional blocks to yield images of size 4x4. Specifically, each convolutional block consists of a 2D transposed convolution layer of stride size 2 followed by a Gaussian error linear unit (GELU) activation function and a 2D convolution of stride size 1. To construct the convolutional block for the decoder part, we use identical layers as in the encoder in reverse order, including the GELU activation function. To obtain the latent vectors, a fully-connected layer with a Tanh activation is used at the end of the encoder network. For the training, an Adam optimizer is used with an initial learning rate of 1e-3 and a learning rate scheduler reducing the learning rate by half every 20 stagnating (non-decreasing reconstruction loss value) epochs. Finally, by incorporating a diffusion model into the trained autoencoder, we showcase the ability of latentPINN to facilitate a more representative PDE parameters sampling. This setup is generally used for all three PDE parameter sets tested below, except for the example where we make some changes to the autoencoder model.

In the second stage, we train a modified PINN that takes in, in addition to the coordinates of the medium, the latent variables that represent the PDE parameters (the velocity models). We also utilize the decoder to reconstruct the velocity model used in the loss function governed by the PDE in equation~\ref{eqt1} (or specifically the factored variation of it) to train the PINN. The reason for using the decoder output instead of the original model, in the loss function of PINN, though they are close, is because the decoder output is a function of the latent vector through the decoder. We utilize a multi-layer perceptron network with an input layer of size 99 ($\textbf{x},\textbf{x}_s$,\textbf{z}, and 96 latent vector variables), an output layer of size 1 ($\hat{\tau}$), 12 hidden layers of size 128 neurons each, and exponential linear unit (ELU) activation function. An Adam optimizer is used with an initial learning rate of 5e-4 and a scheduler to reduce the learning rate by half every 200 stagnating (non-decreasing PDE loss values) epochs. We randomly select 9830 collocation points as the training samples. We train the network with 100 velocity models for 10k epochs with a batch size of 163 points. We randomly sample these velocities (by their latent vectors) such that every iteration corresponds to solving for a specific velocity field. Again, this setup for the latentPINN is generally used to solve for the eikonal equation for all three tests below, except we make some changes for the last test.

 All experiments were performed on a single NVIDIA Quadro RTX 8000 GPU card utilizing the PyTorch machine learning framework \cite[]{paszke2019pytorch}.
 We will assess the accuracy of the learned traveltime functions by comparing their predictions with the numerical solutions. We also test the accuracy of the PINN solution by comparing the reconstructed velocity models (plugging in the predicted PDE solutions $T$ into equation \ref{eqt1}) to the output of the decoder $\hat{V}$.

\begin{figure}[!ht]
\begin{subfigure}{.5\textwidth}
    \centering
    \includegraphics[width=0.6198\linewidth]{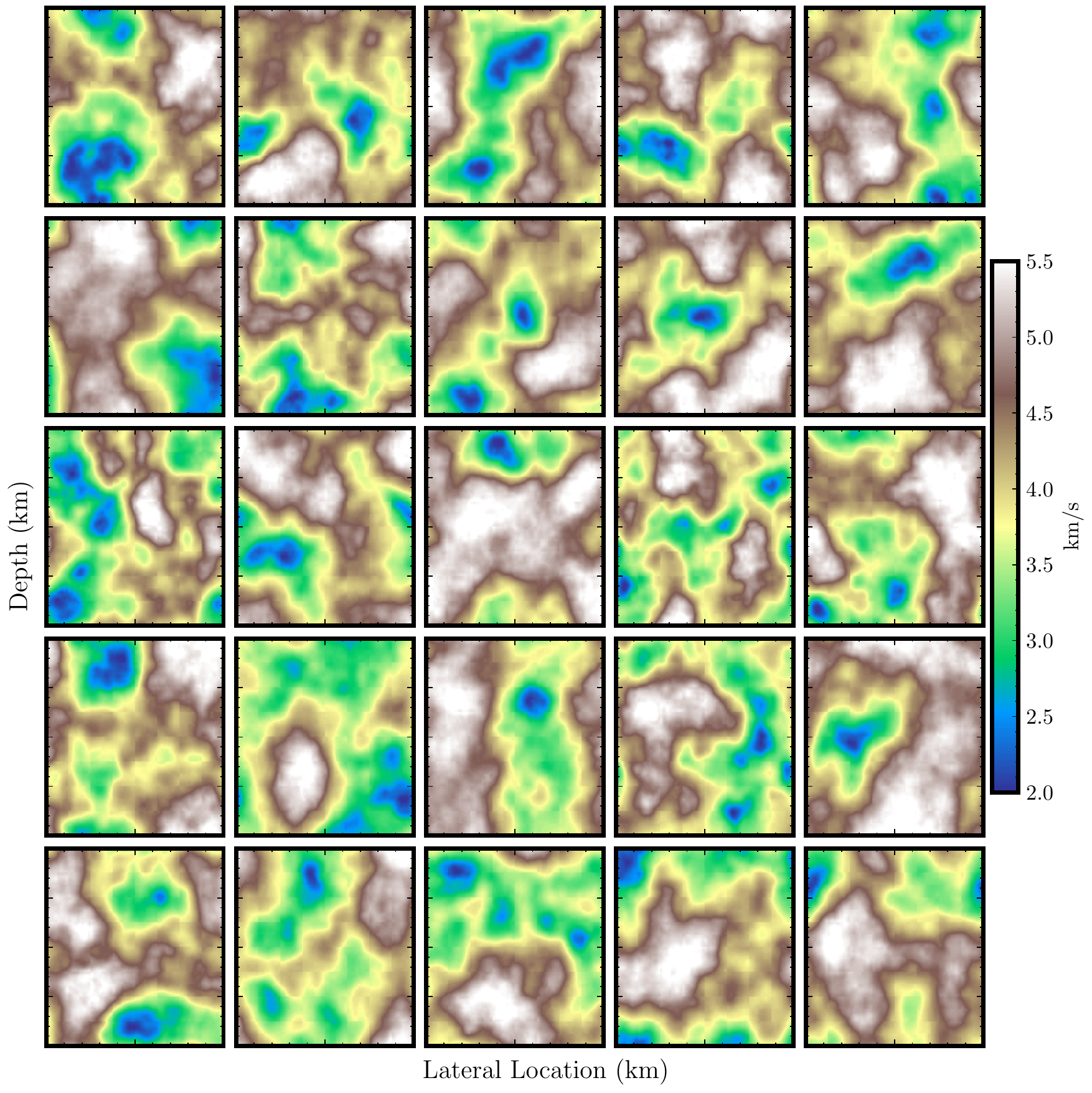}
    \caption{}
    \label{recona}
\end{subfigure}
\begin{subfigure}{.5\textwidth}
    \centering
    \includegraphics[width=0.6198\linewidth]{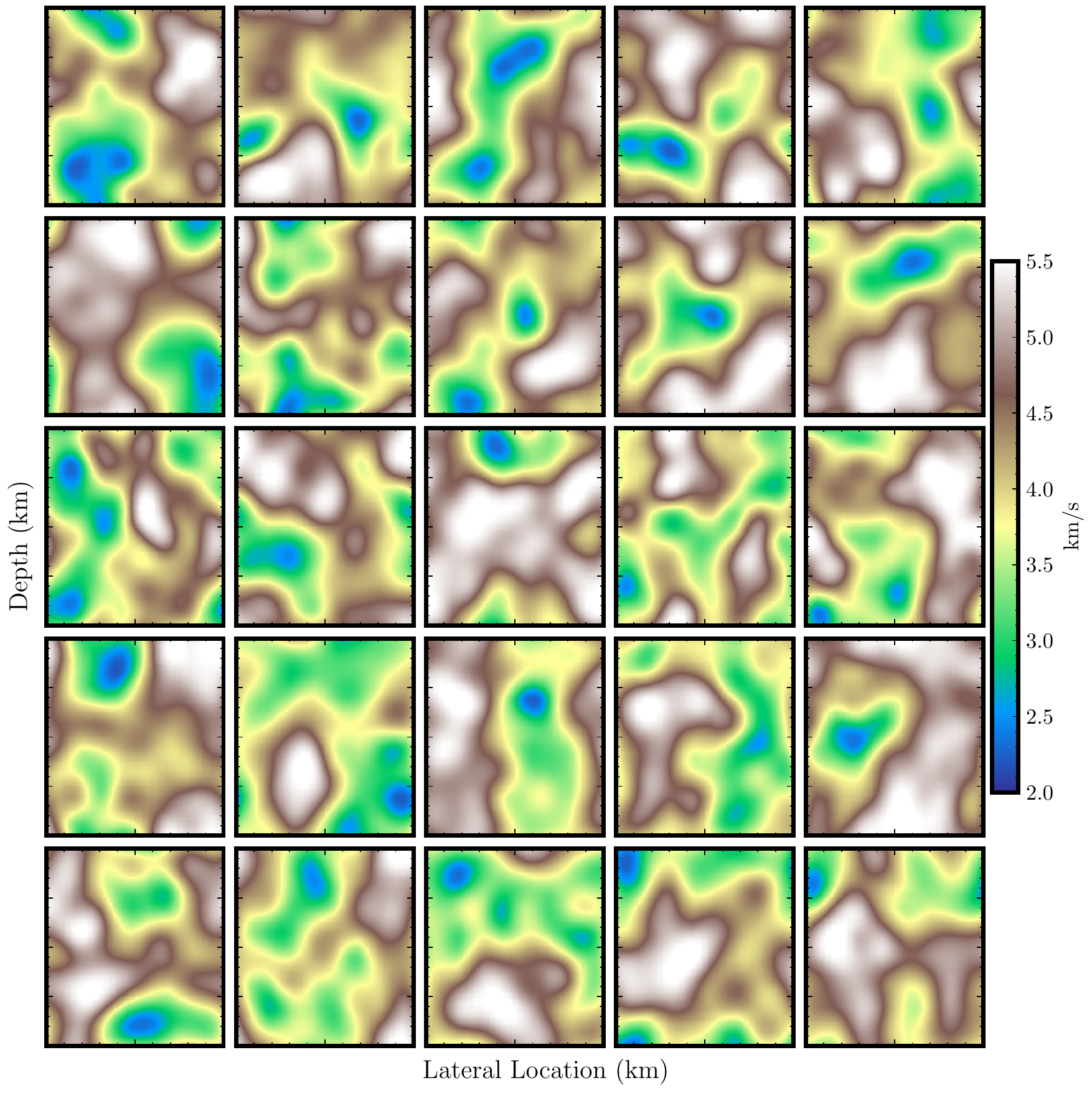}
    \caption{}
    \label{reconb}
\end{subfigure}
\begin{subfigure}{.5\textwidth}
    \centering
    \includegraphics[width=0.6198\linewidth]{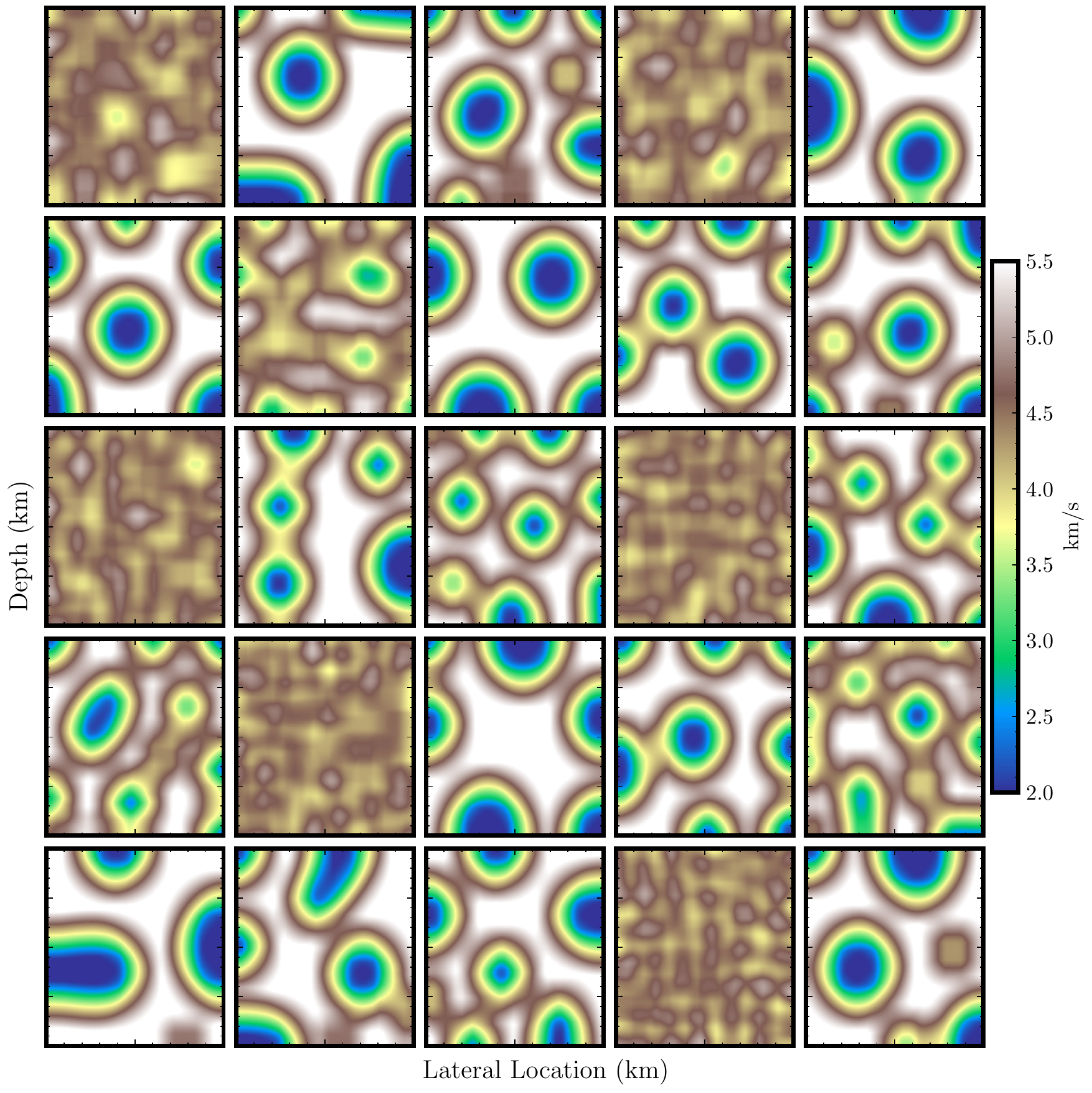}
    \caption{}
    \label{reconc}
\end{subfigure}
\begin{subfigure}{.5\textwidth}
    \centering
    \includegraphics[width=0.6198\linewidth]{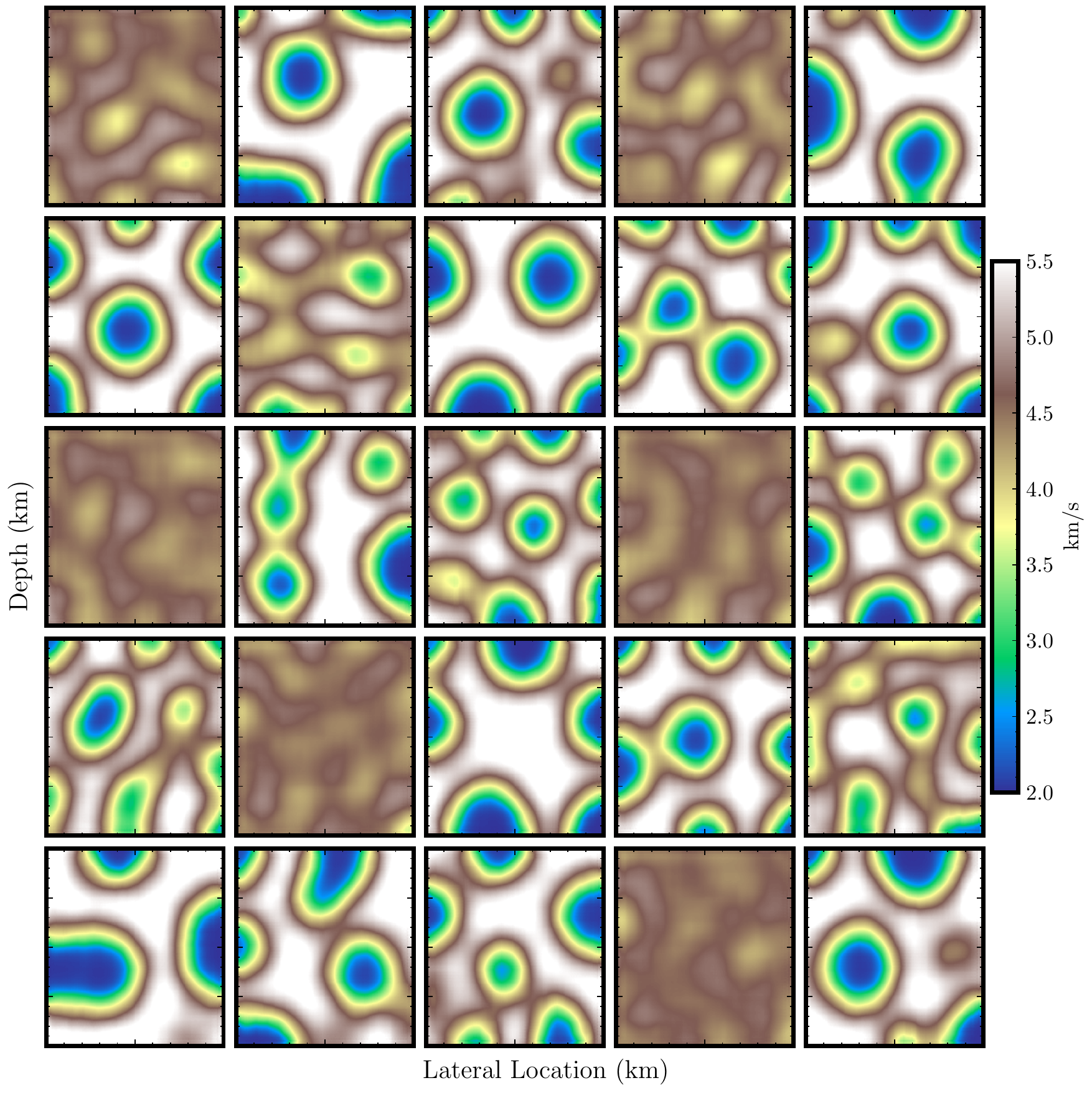}
    \caption{}
    \label{recond}
\end{subfigure}
\begin{subfigure}{.5\textwidth}
    \centering
    \includegraphics[width=0.6198\linewidth]{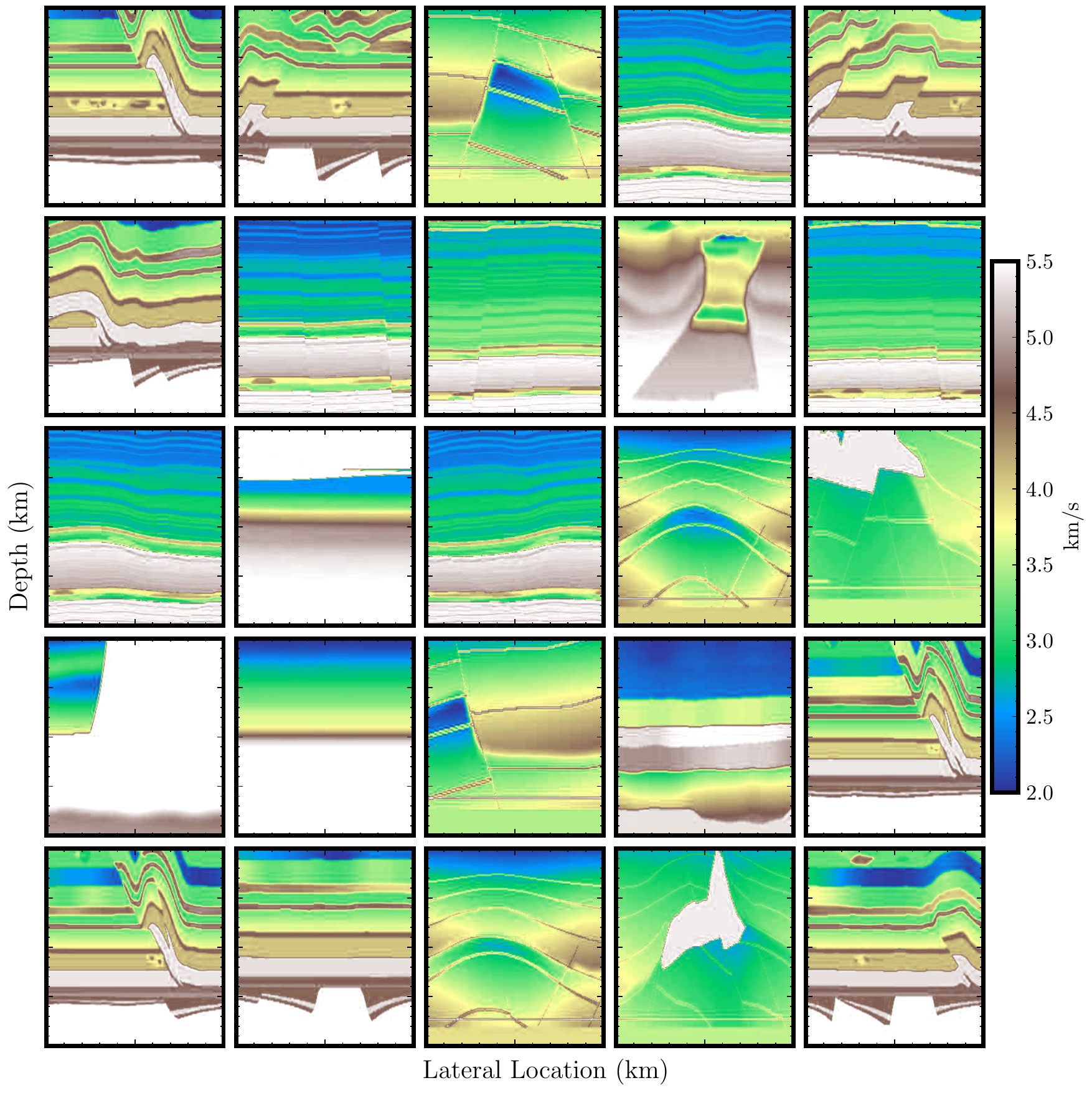}
    \caption{}
    \label{recone}
\end{subfigure}
\begin{subfigure}{.5\textwidth}
    \centering
    \includegraphics[width=.6\linewidth]{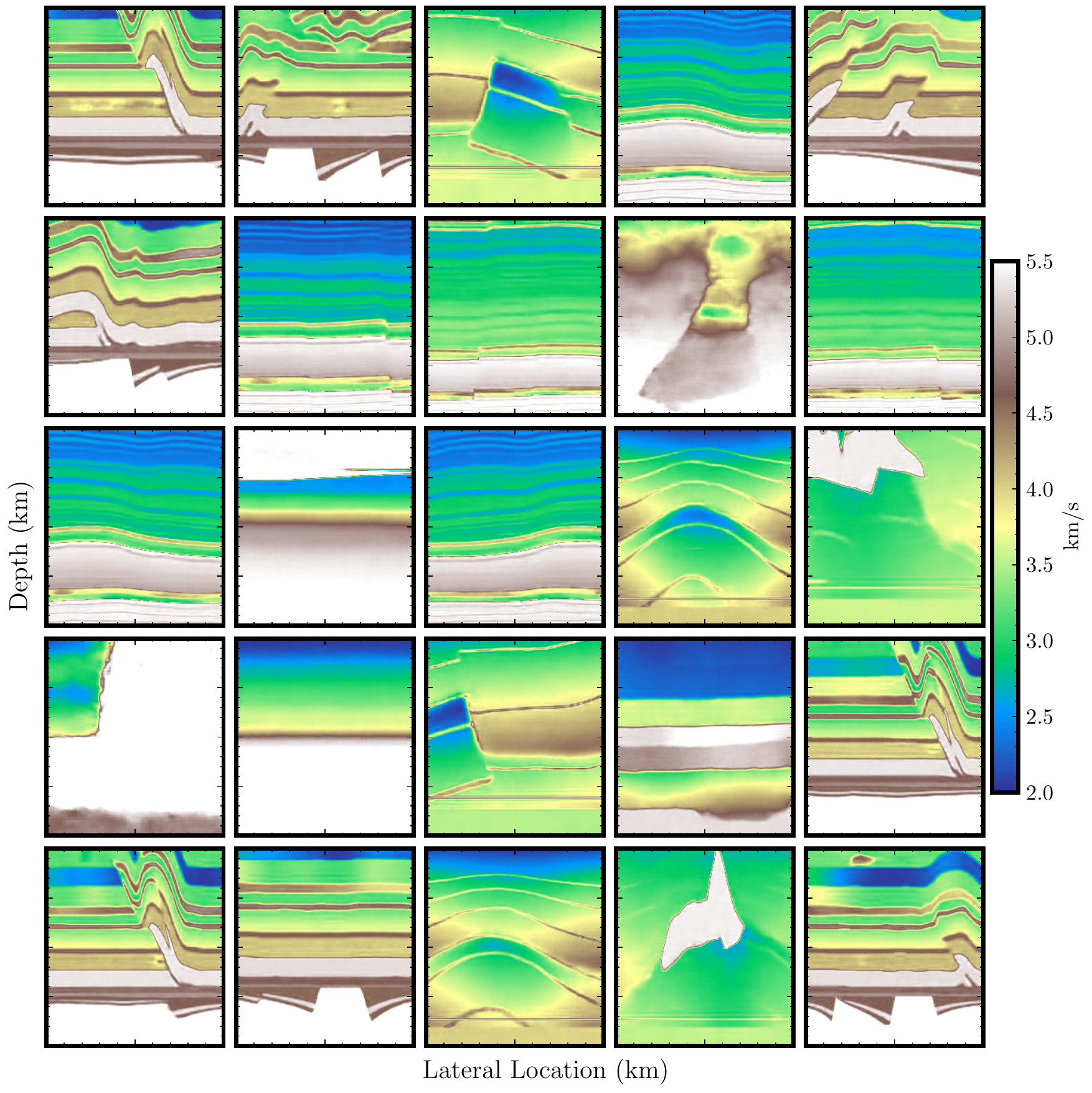}
    \caption{}
    \label{reconf}
\end{subfigure}
\caption{
Reconstruction analysis on three different datasets; Gaussian random fields (a and b), Mechanical MNIST examples (c and d), and realistic Earth models (e and f). The left column includes the original samples drawn from the testing set (a,c,e), and the right column includes the reconstructed samples (b,d,f).}
\label{recon}
\end{figure}

\begin{figure}
\begin{subfigure}{\textwidth}
    \centering
    \includegraphics[width=0.45\linewidth]{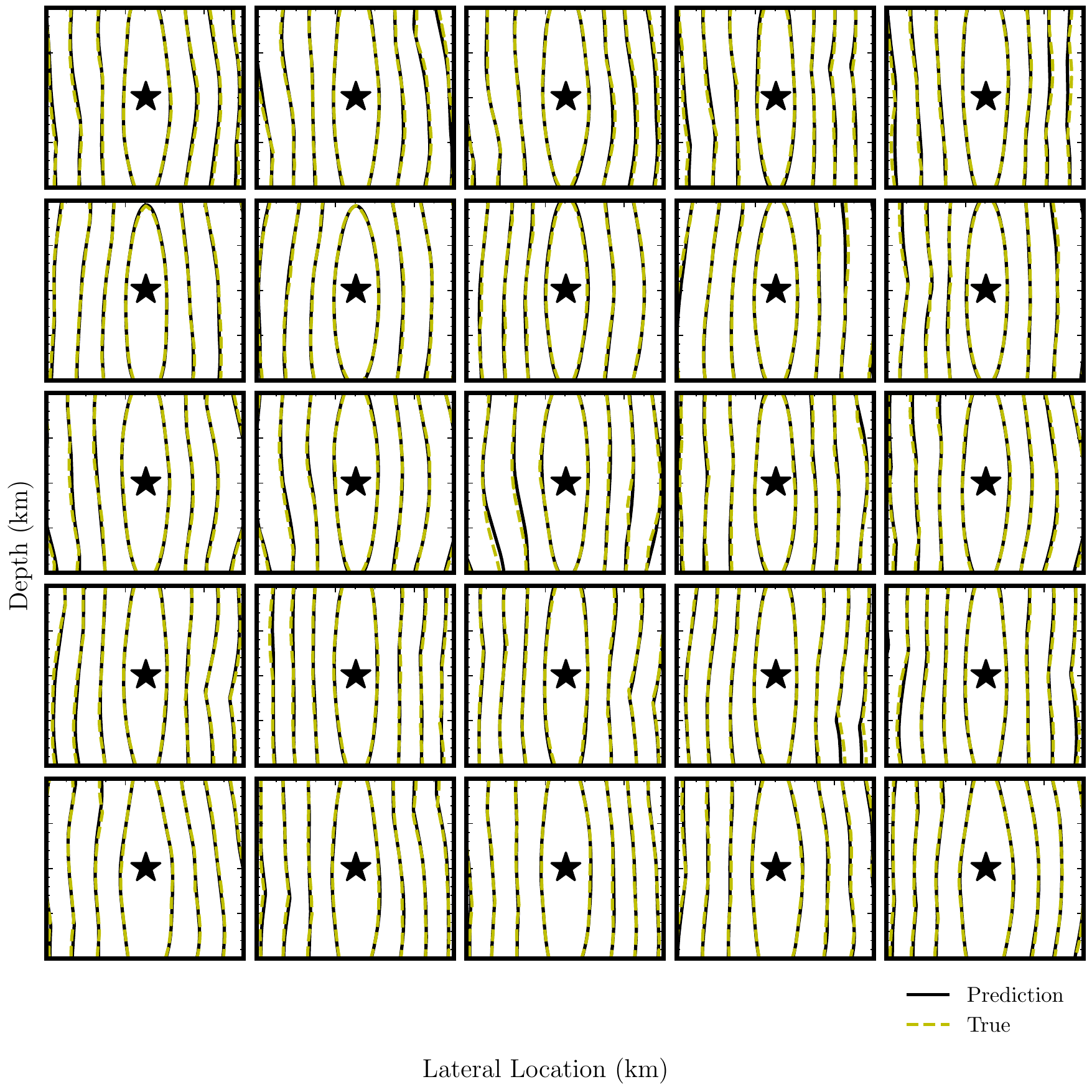}
    \caption{}
    \label{solutionsa}
\end{subfigure}
\begin{subfigure}{\textwidth}
    \centering
    \includegraphics[width=0.45\linewidth]{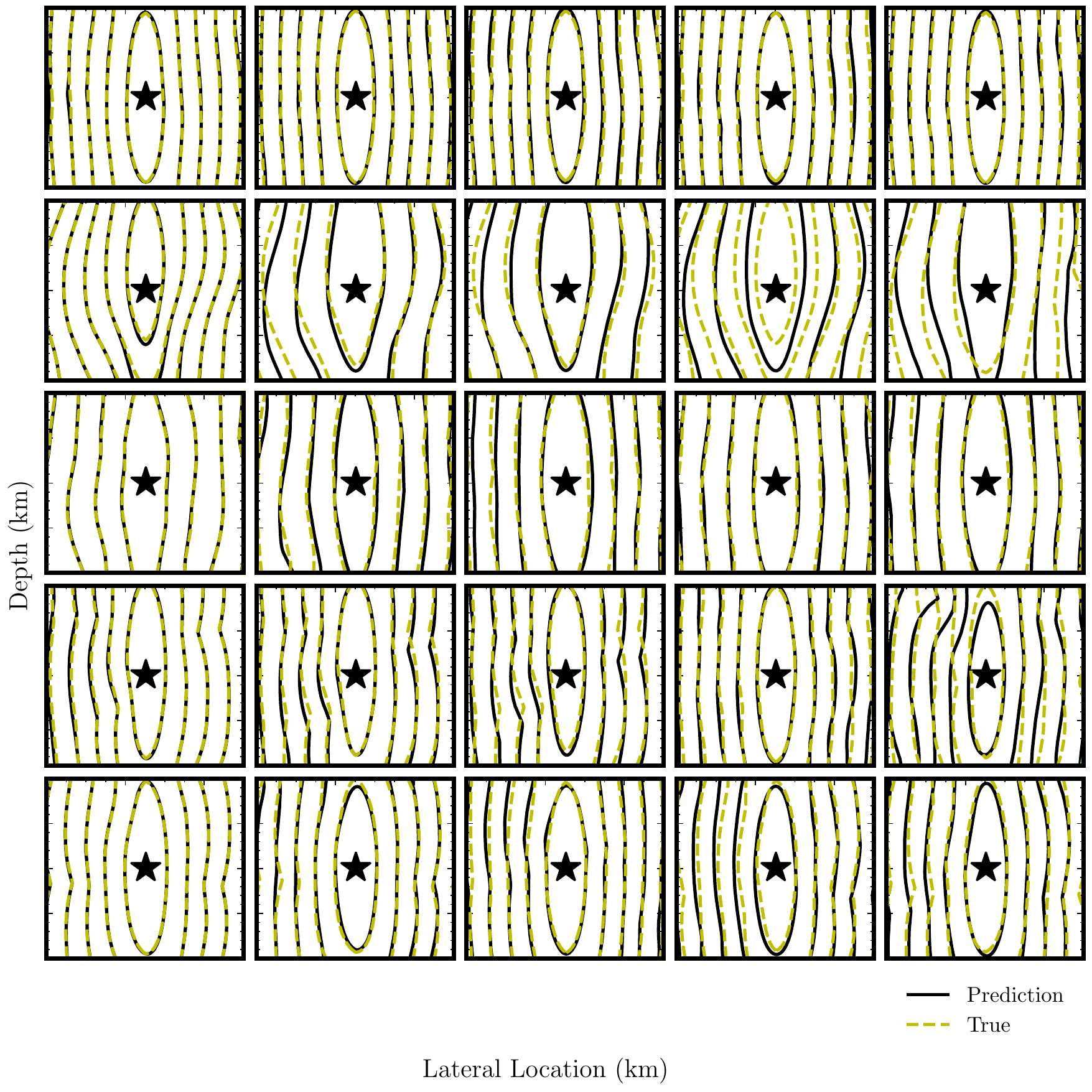}
    \caption{}
    \label{solutionsb}
\end{subfigure}
\begin{subfigure}{\textwidth}
    \centering
    \includegraphics[width=0.45\linewidth]{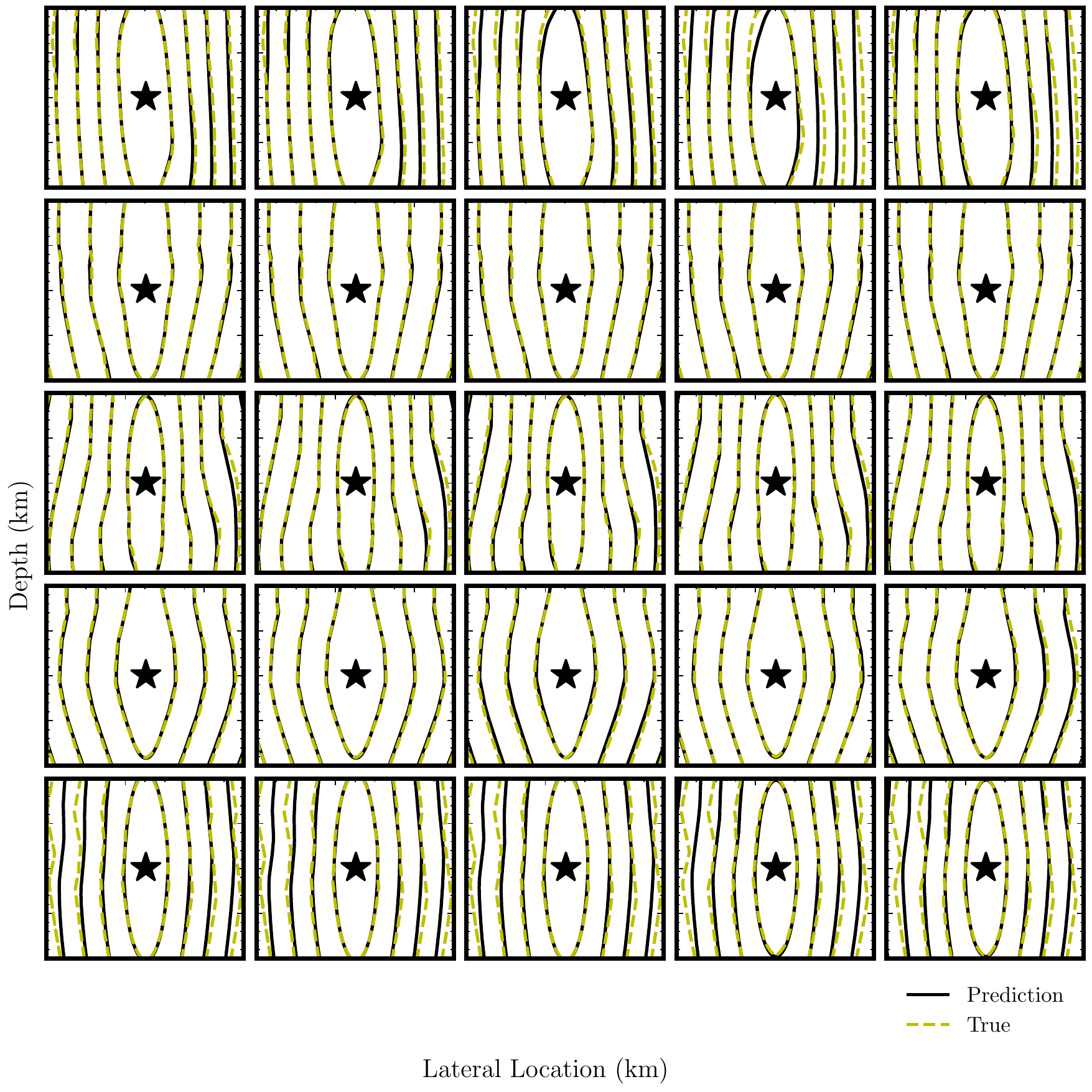}
    \caption{}
    \label{solutionsc}
\end{subfigure}
\caption{
Predicted PDE solutions (traveltimes, solid black lines), compared with conventional solvers (dashed yellow lines) from the three datasets; a) Gaussian random fields, b) Mechanical MNIST, and c) realistic Earth models. The leftmost column of each plot corresponds to solutions from a seen training PINN sample (velocity model), while the rest of the columns correspond to samples from the unseen testing set.}
\label{solutions}
\end{figure}

\begin{figure}[!ht]
\begin{subfigure}{.5\textwidth}
    \centering
    \includegraphics[width=0.6198\linewidth]{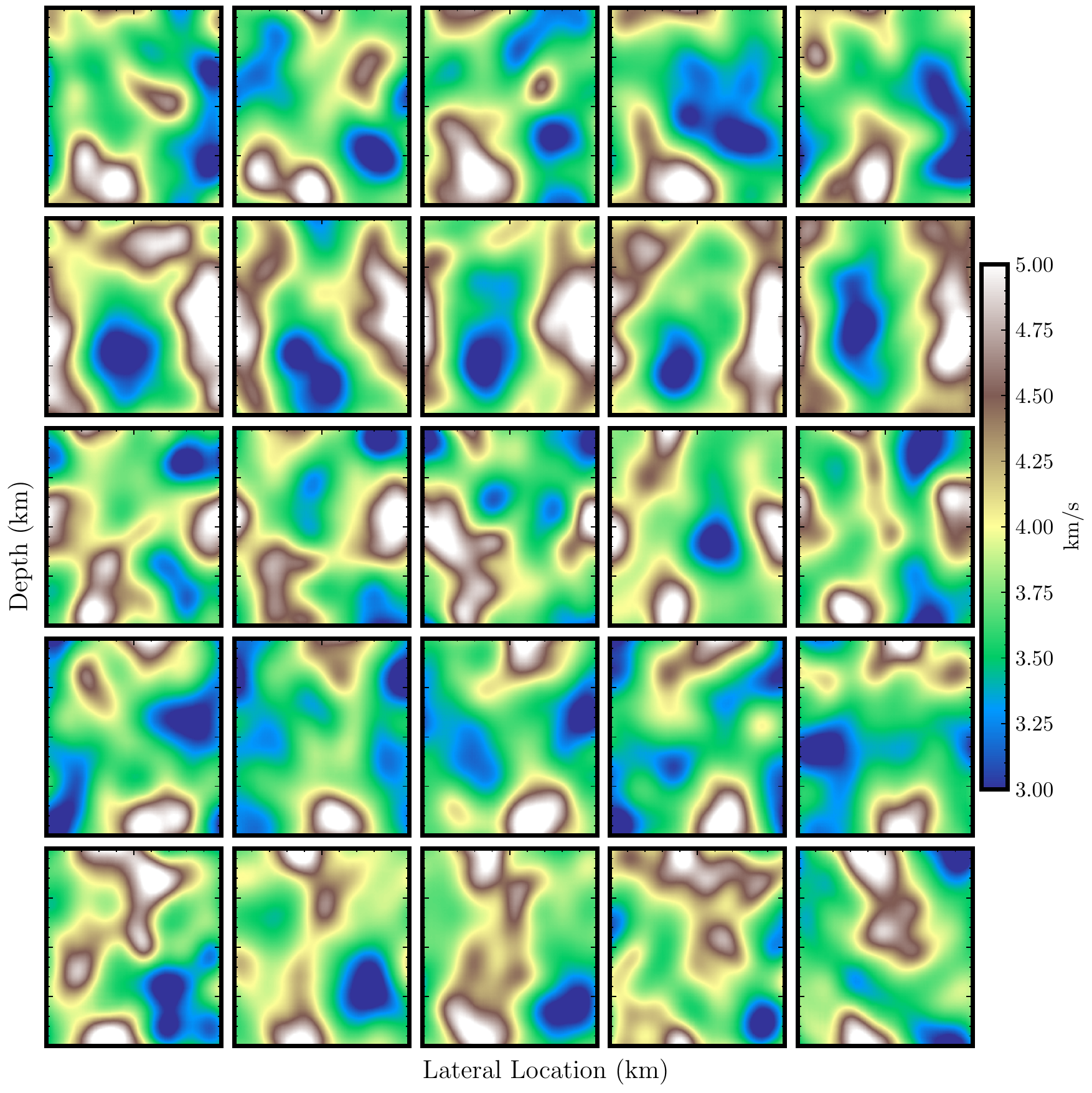}
    \caption{}
    \label{pinnsa}
\end{subfigure}
\begin{subfigure}{.5\textwidth}
    \centering
    \includegraphics[width=0.6198\linewidth]{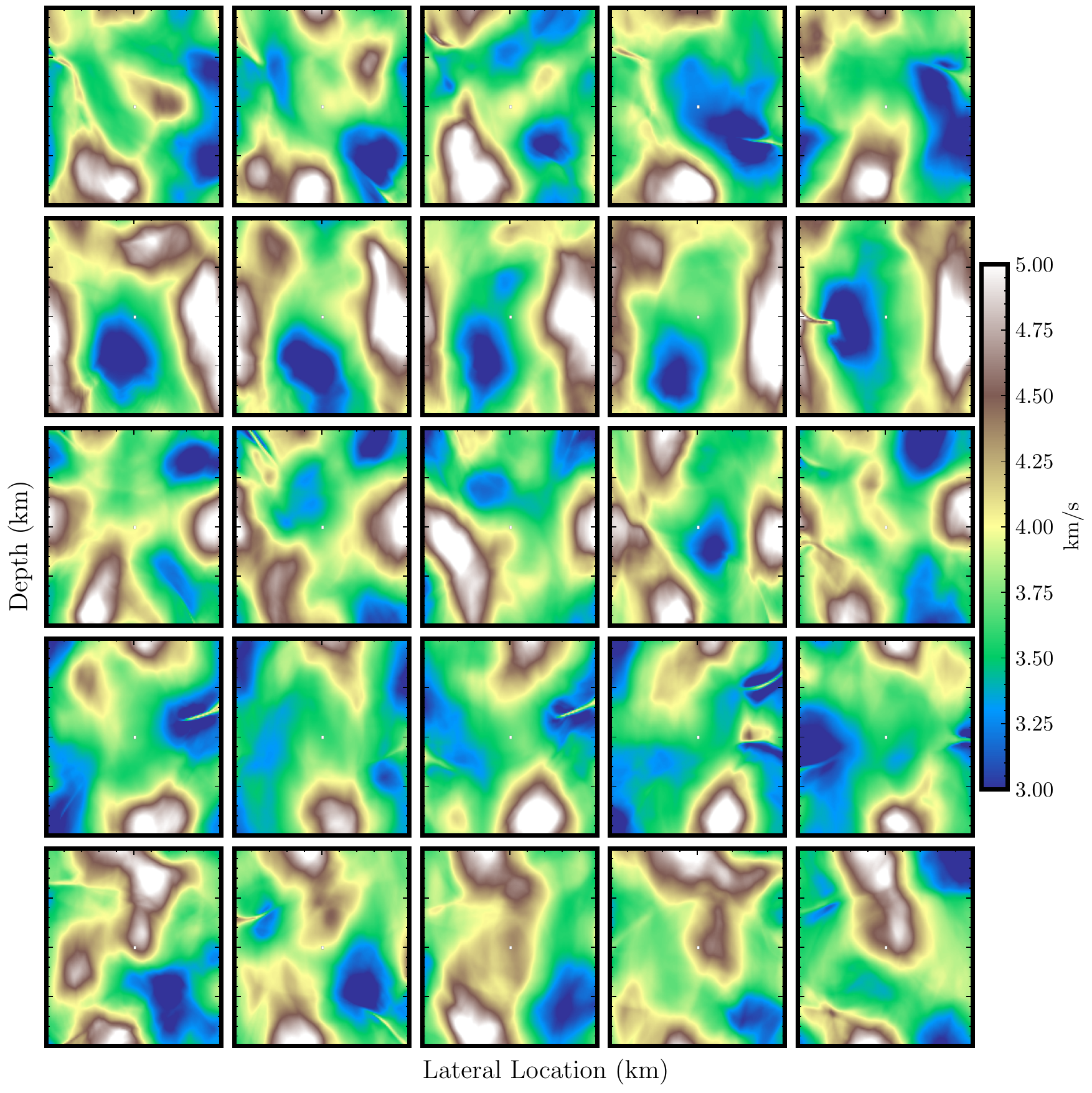}
    \caption{}
    \label{pinnsb}
\end{subfigure}
\begin{subfigure}{.5\textwidth}
    \centering
    \includegraphics[width=0.6198\linewidth]{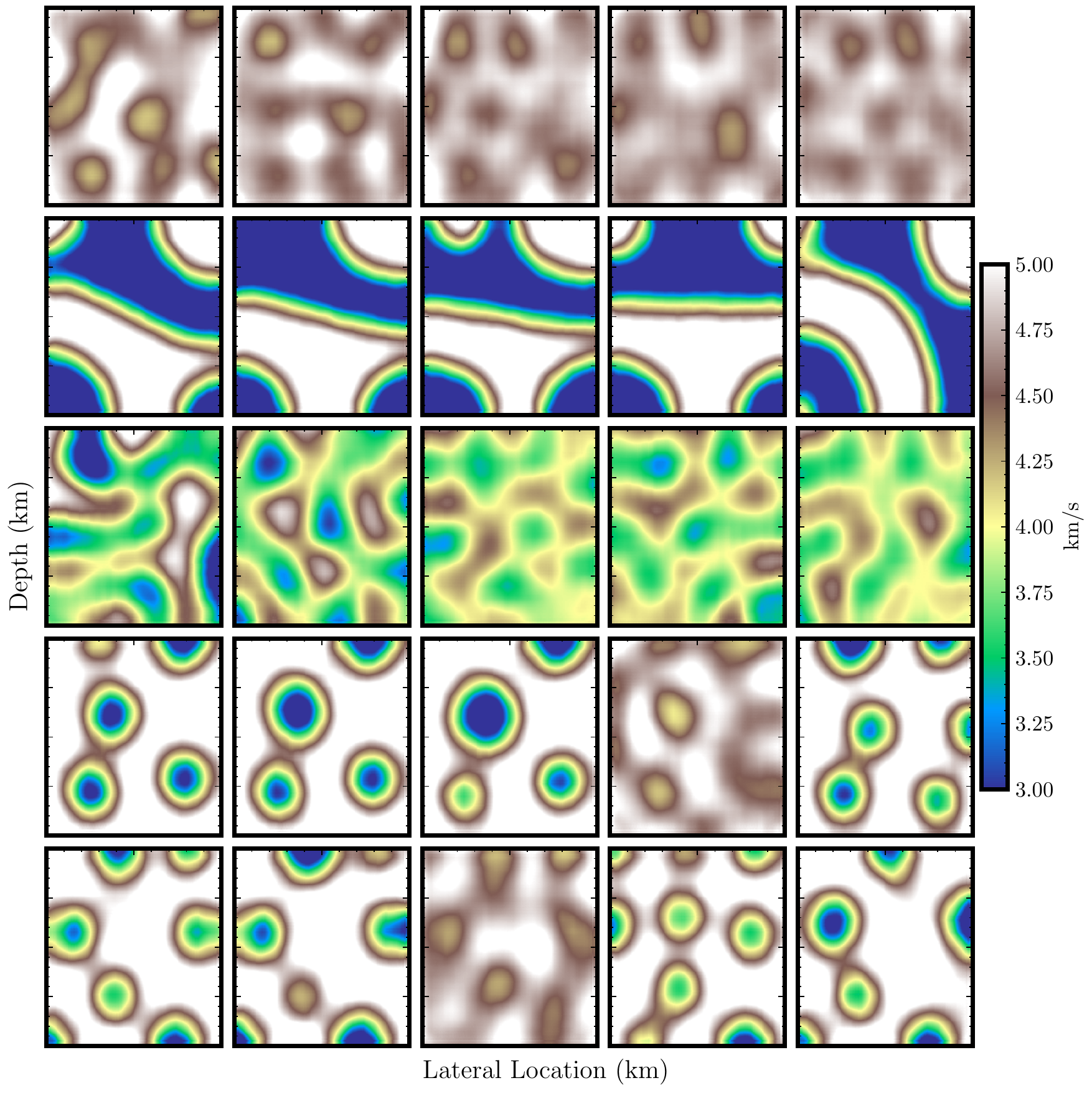}
    \caption{}
    \label{pinnsc}
\end{subfigure}
\begin{subfigure}{.5\textwidth}
    \centering
    \includegraphics[width=0.6198\linewidth]{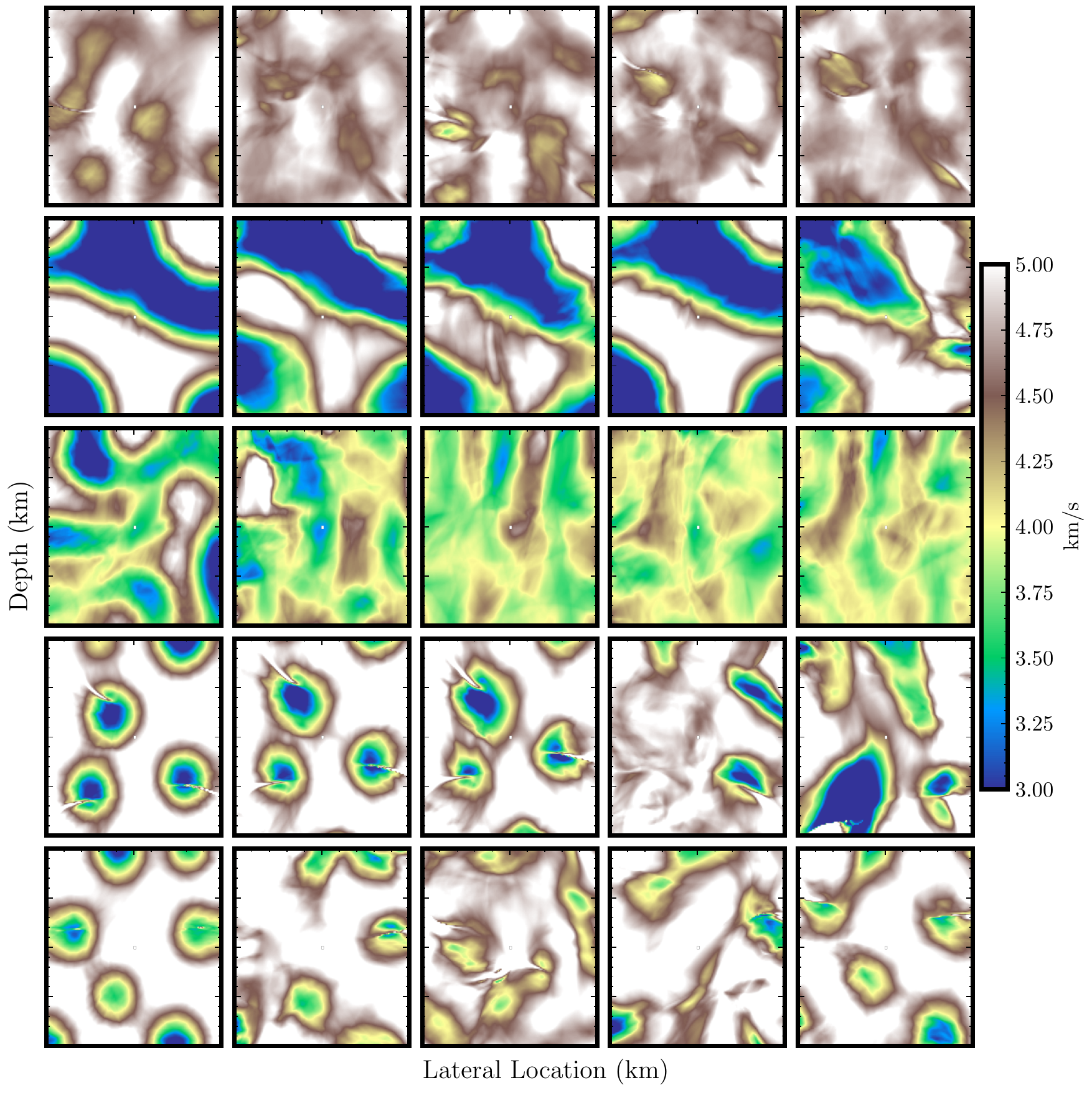}
    \caption{}
    \label{pinnsd}
\end{subfigure}
\begin{subfigure}{.5\textwidth}
    \centering
    \includegraphics[width=0.6198\linewidth]{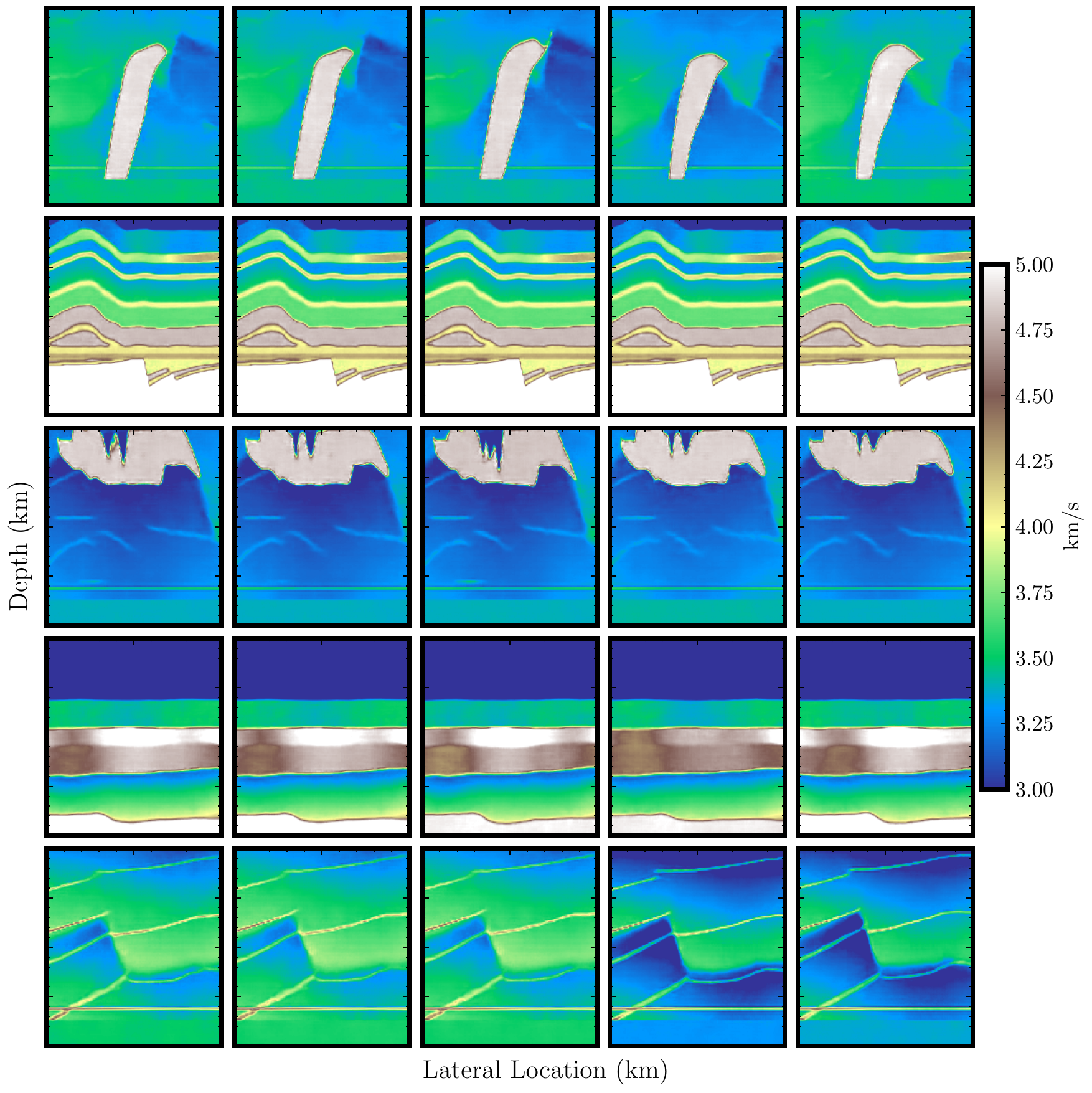}
    \caption{}
    \label{pinnse}
\end{subfigure}
\begin{subfigure}{.5\textwidth}
    \centering
    \includegraphics[width=.6\linewidth]{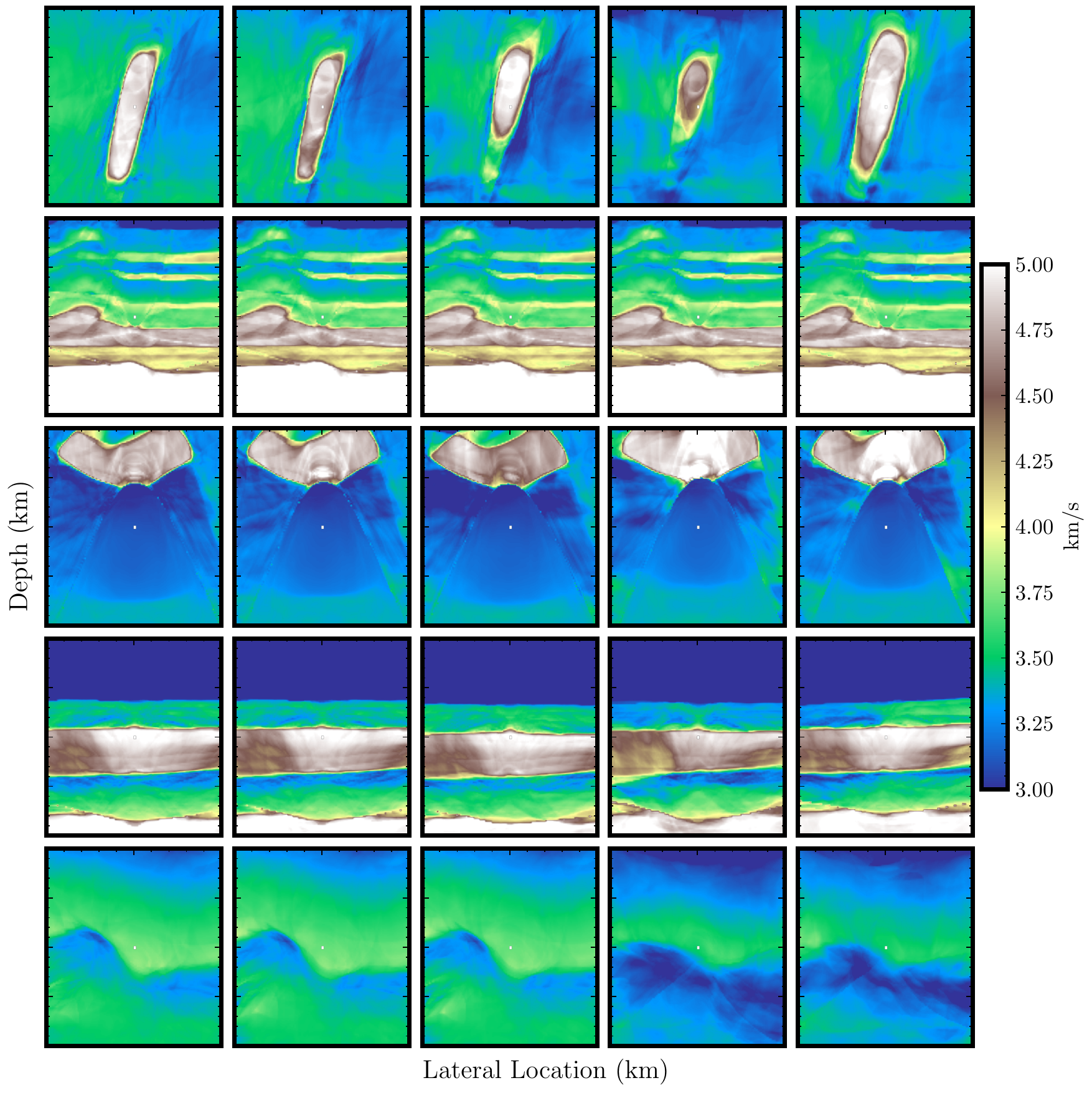}
    \caption{}
    \label{pinnsf}
\end{subfigure}
\caption{
Accuracy analysis on three different datasets from the reconstructed PDE parameters (using equation \ref{eqt1}). The left column plots show the original samples drawn (a,c,e), and the right column plots include the reconstructed samples (b,d,f). The left most samples from each plot correspond to velocity models from the training set; the rest of the columns include results for unseen PDE parameters (from the testing set).}
\label{pinns}
\end{figure}

\begin{figure}[!ht]
\begin{subfigure}{.33\textwidth}
    \centering
    \includegraphics[width=0.939\linewidth]{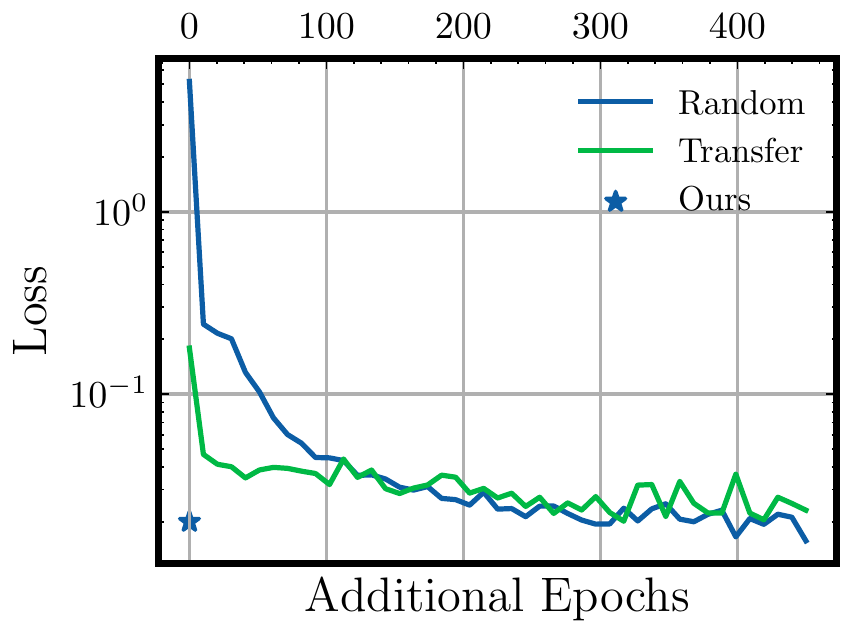}
    \caption{}
    \label{ooda}
\end{subfigure}
\begin{subfigure}{.33\textwidth}
    \centering
    \includegraphics[width=0.939\linewidth]{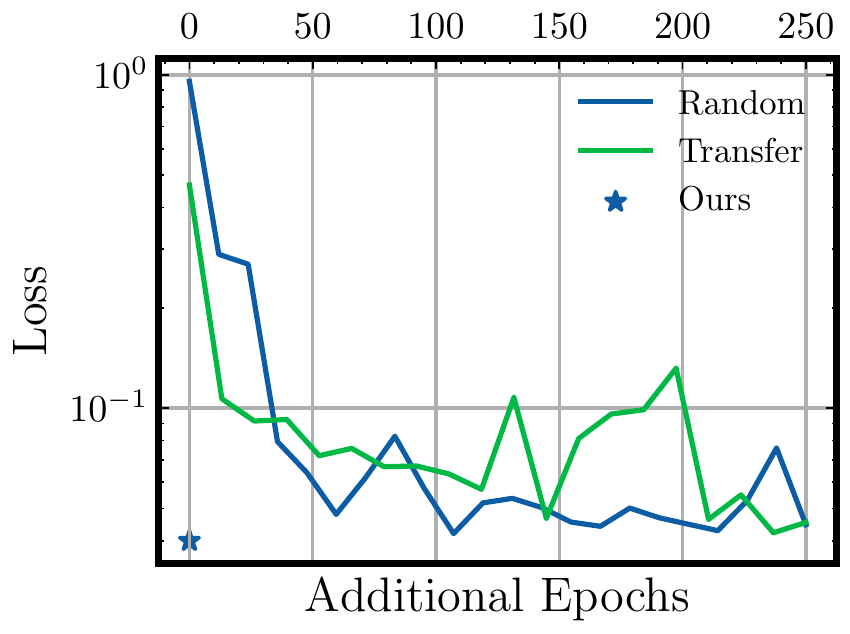}
    \caption{}
    \label{oodb}
\end{subfigure}
\begin{subfigure}{.33\textwidth}
    \centering
    \includegraphics[width=0.939\linewidth]{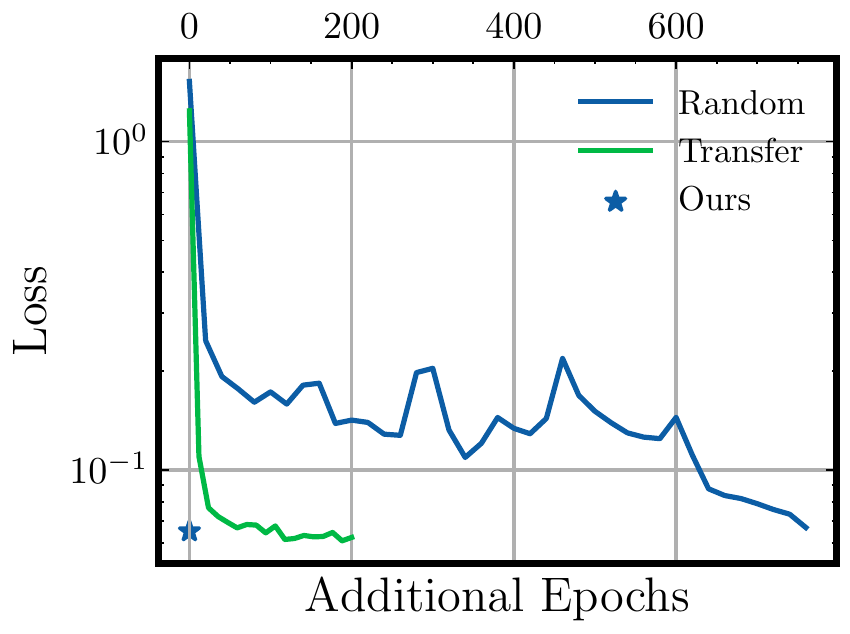}
    \caption{}
    \label{oodc}
\end{subfigure}
\begin{subfigure}{.33\textwidth}
    \centering
    \includegraphics[width=0.939\linewidth]{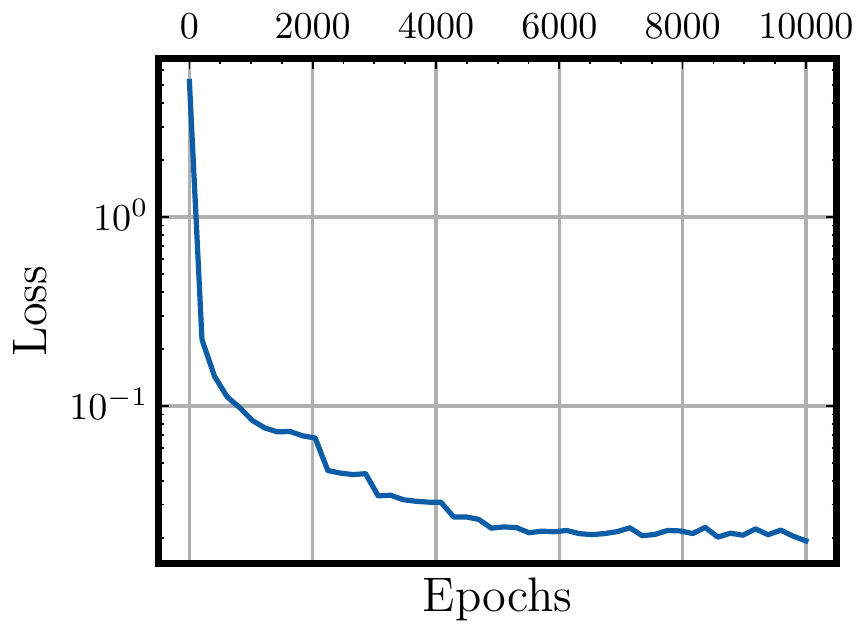}
    \caption{}
    \label{oodd}
\end{subfigure}
\begin{subfigure}{.33\textwidth}
    \centering
    \includegraphics[width=0.939\linewidth]{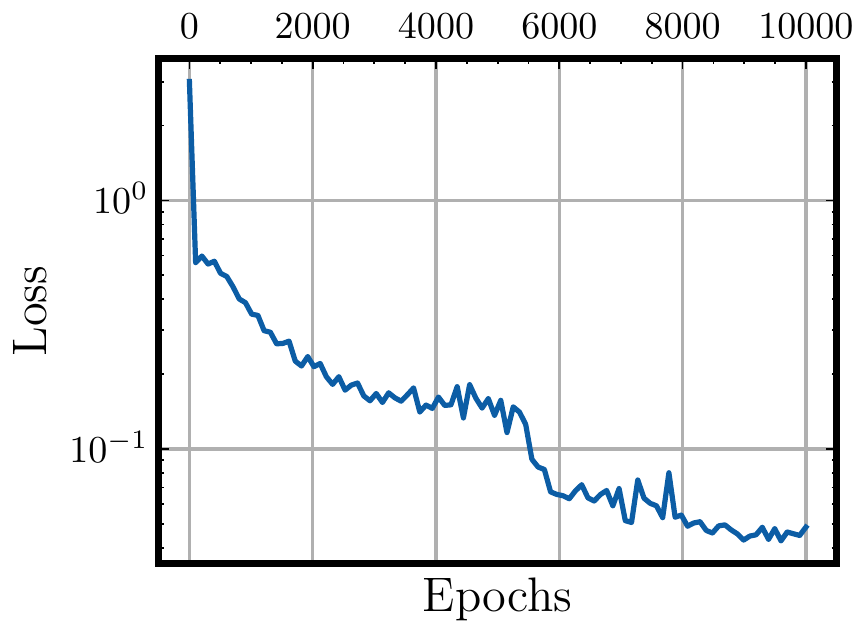}
    \caption{}
    \label{oode}
\end{subfigure}
\begin{subfigure}{.33\textwidth}
    \centering
    \includegraphics[width=0.939\linewidth]{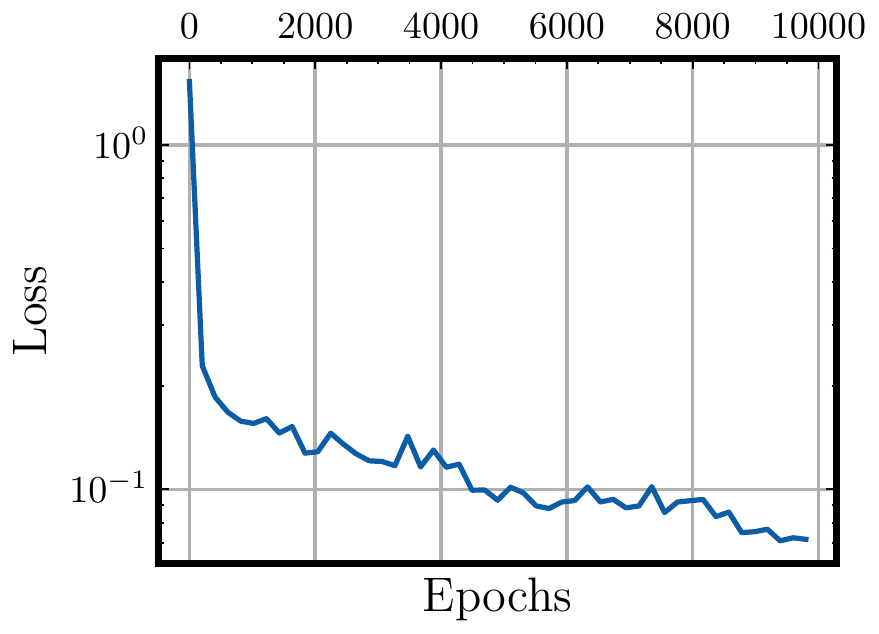}
    \caption{}
    \label{oodf}
\end{subfigure}
\caption{
Loss curves comparison between our approach and several training mechanisms. The top row (a to c) compares the required additional epochs to train for a new PDE parameter between the trained latentPINN (blue stars) and conventional PINN training from randomly initialized weights (blue lines) and performing transfer learning (green lines). The bottom row (d to f) details the training cost to train our latentPINN model for 100 randomly chosen velocity models (PDE parameters) from the training set used previously in the autoencoder training.}
\label{ood}
\end{figure}

\subsection{Numerical Results}

The goal here is to test the ability of the latent implementation to generalize on test PDE parameter examples not used in the training of latentPINN. To do so, we consider three sets of PDE parameters (phase velocity model distributions): 1) A set drawn from Gaussian random fields, 2) a set drawn from a Mechanical MNIST distribution, and finally, 3) a set related to examples of layering inside the Earth.

\subsubsection{Gaussian Random Fields}

In the first scenario, we use Gaussian random fields (GRFs) to generate the set of PDE parameters for learning the latent space and training the latentPINN. The isotropic 2D spatial Gaussian random field is given by
\begin{equation}
\boldsymbol{x} = \mathcal{F}^{-1} \left(C \cdot \mathrm{n}^2 \cdot \sqrt{2} \cdot \tau^{\frac{1}{2}(2\alpha-2)} \cdot \left(\frac{4\pi^2(k_x^2+k_x'^2)+\tau^2}{2}\right)^{-\alpha/2} \right), \quad \boldsymbol{x} \in \mathbb{R}^2, \quad C \in \mathcal{N}(0, \text{I}),
\end{equation}
where an inverse Fourier transform ($\mathcal{F}^{-1}$) is performed on a correlation function parameterized by the pixel dimension $n$, spatial wavenumbers $k_x \in [0,\frac{\text{n}}{2}]$, and parameters controlling the variance, $\tau$ and $\alpha$ in its exponent argument ($\sigma = \tau^{0.5*(2*\alpha - \text{n})}$ where $\sigma$ is the intended variance). Specifically, we generated 50k image samples ($\sim \mathcal{N}(0, \text{I})$) of dimension 128x128 with $\tau=7$, $\alpha=2.5$, and a periodic boundary condition.  Out of the 50k three-channel image samples, 39k were used for the training set, 1k for validation, and the rest for testing. These near zero-mean images that range between -1 and 1 are then converted into velocity fields by first adding a constant of 2 km/s followed by a multiplication by a factor of 2. Thus, the resulting velocity range is [2,6] km/s. Figure \ref{recona} shows samples out of the 50k generated velocity fields. 

 For this PDE parameter distribution, we train an autoencoder to learn the corresponding latent representation for 30 epochs. Though slightly smoothed out, we can see the well-reconstructed images from the test set in Figure \ref{reconb}, reflecting that it learned the compressed latent representation. Out of the 39k training samples, 100 are used during the second stage, the main PINN training.
Shown in the left-most column of each plot in Figures \ref{pinns} and \ref{solutions} are velocities (out of the 100 samples) used during the PINNs training, while the rest correspond to the unseen (testing) velocities. Going from left to right, for each row, we order the testing samples based on an increasing value of point-wise distance in the latent space. With such an order, we expect to have more dissimilar images with increasing point-wise distance. We can see in Figures \ref{pinnsb} and \ref{solutionsa} that the trained PINNs generalize to new velocities well. It is noteworthy to mention that the solutions for the velocities from the test set (of the Gaussian Random field set) are without any additional training, unlike for vanilla PINNs (Figure \ref{ooda}).

\subsubsection{Mechanical MNIST}

Next, we consider a slightly more complex dataset in the form of the solutions to the Cahn-Hilliard equation \cite[]{cahn1958free}. The fourth-order parabolic PDE details the phase separation dynamics of a two-phase mixture. In our case, we utilize such a dataset to understand the behavior of our approach in the face of a sparser (in pixel values) data distribution. We consider the use of a smoothed version of the Mechanical MNIST dataset \cite{https://open.bu.edu/handle/2144/43971}. We consider training, validation, and testing sets of sizes 15784, 1000, and 9420, respectively. The three-channel 128x128 are converted into velocity fields using the same addition and multiplication constants as in the previous case ($=2 (x+2)$ km/s, where $x$ is the Mechanical MNIST pixel value, between -1 and 1). Figure \ref{reconc} illustrates samples drawn from the training set.

Using the same training configuration and autoencoder model as in the previous GRF case, we obtain after training well-reconstructed image samples from the testing set in Figure \ref{recond}. We assess the learned latent vectors after the main PINNs training. For this training, we utilize a similar training configuration as in the previous case with a slight modification. Considering the complexity of this set (more variance), we double the hidden layers (from 12 to 24). We train the network for 10k epochs of batch size 163 points. We test the accuracy of the predicted PDE solutions on test samples given by the four rightmost columns in Figures \ref{pinnsc} (input) and \ref{pinnsd}) (reconstruction). We utilize the trained encoder to provide latent vectors for the training of the PINN, and we use the decoder to provide the velocity model used in the PDE loss function. There is no clear distinction in the efficiency between randomly initialized and transfer learning training (Figure \ref{oodb}). This might be related to the inherent limitations of the Eikonal equation in dealing with such complex PDE parameters set (velocity models). As we are dealing with sharper boundaries between two phases in the Mechanical MNIST dataset, the PINNs training becomes slightly more challenging than for the previous smoother spatial Gaussian random fields. Nonetheless, our trained latentPINN still provides new PDE solutions given new PDE parameters without additional PINN training.

\subsubsection{Realistic Earth Models}

Finally, we consider a PDE parameter distribution for applications in traveltime modeling of the Earth subsurface in the form of realistic synthetic subsurface velocity models (Figure \ref{recone}). We compile patches of images from several 2D velocity images \cite[]{billette20052004,martin2006marmousi2} and 3D velocity cubes \cite[]{Aminzadeh_1997_DSO,naranjo2011survey} resulting in a 70,522 256x256 three-channel images. The dataset entails a wide range of geological settings ranging from highly faulted models to anomalous geological bodies (e.g., salts and low-velocity layers). We utilize a 90-5-5 percentage split to generate the training-validation-testing datasets. Figures \ref{recone} and \ref{pinnse} illustrate samples drawn from the datasets.

For the first stage of the latentPINN training, we modify the encoder-decoder network to perform down- and upsampling of the higher-resolution images. Specifically, each convolutional block now consists of a 2D transposed convolution layer of stride size 4 followed by a GELU layer and a 2D convolution of stride size 1. We utilize two of these blocks in the encoder and decoder (reversing the order within the encoder convolution block) network. The same optimizer, latent vector size, and learning rate scheduler (as in the previous two cases) are used for the training. We train this model for 45 epochs with a batch size of 256 samples. The well-reconstructed images shown in Figure \ref{reconf} demonstrate the well-represented data distribution from the compressed latent vectors. The quality of these vectors can be further attested from the PINNs training results shown in Figure \ref{pinnsf}. For this training, we utilize almost identical training dynamics (as in the Mechanical MNIST) with a slight change in the initial learning rate. We start the training with a 3e-4 learning rate. Our trained latentPINN manages to generalize to the new testing samples (the four rightmost columns in Figures \ref{pinnse} and \ref{pinnsf}). The subtle changes (from left to right) can be captured by the trained PINN without additional training. Figure \ref{oodc} further details the saving in training costs that would be otherwise needed in conventional approaches. Here, we only have to contend with the overhead cost of training the latentPINN.

\subsection{Functional Data Representations}

One notable feature of our proposed latentPINN framework is its readily extensible to sample functions over the continuous domain (e.g., PDE parameters), termed \emph{functional data} representation. In such a task, we aim to perform infinite-dimensional sampling of the functional data accommodated by a neural network function. Such representation is an important feature for a wide range of PDE applications. For example, being able to perform this sampling benefits application related but not limited to imaging the Earth's interior \cite[]{yang2021seismic}, weather forecasting \cite[]{pathak2022fourcastnet}, and fluid flow simulation \cite[]{wen2022u}.

\begin{figure}[!ht]
\begin{subfigure}{.33\textwidth}
    \centering
    \includegraphics[width=0.939\linewidth]{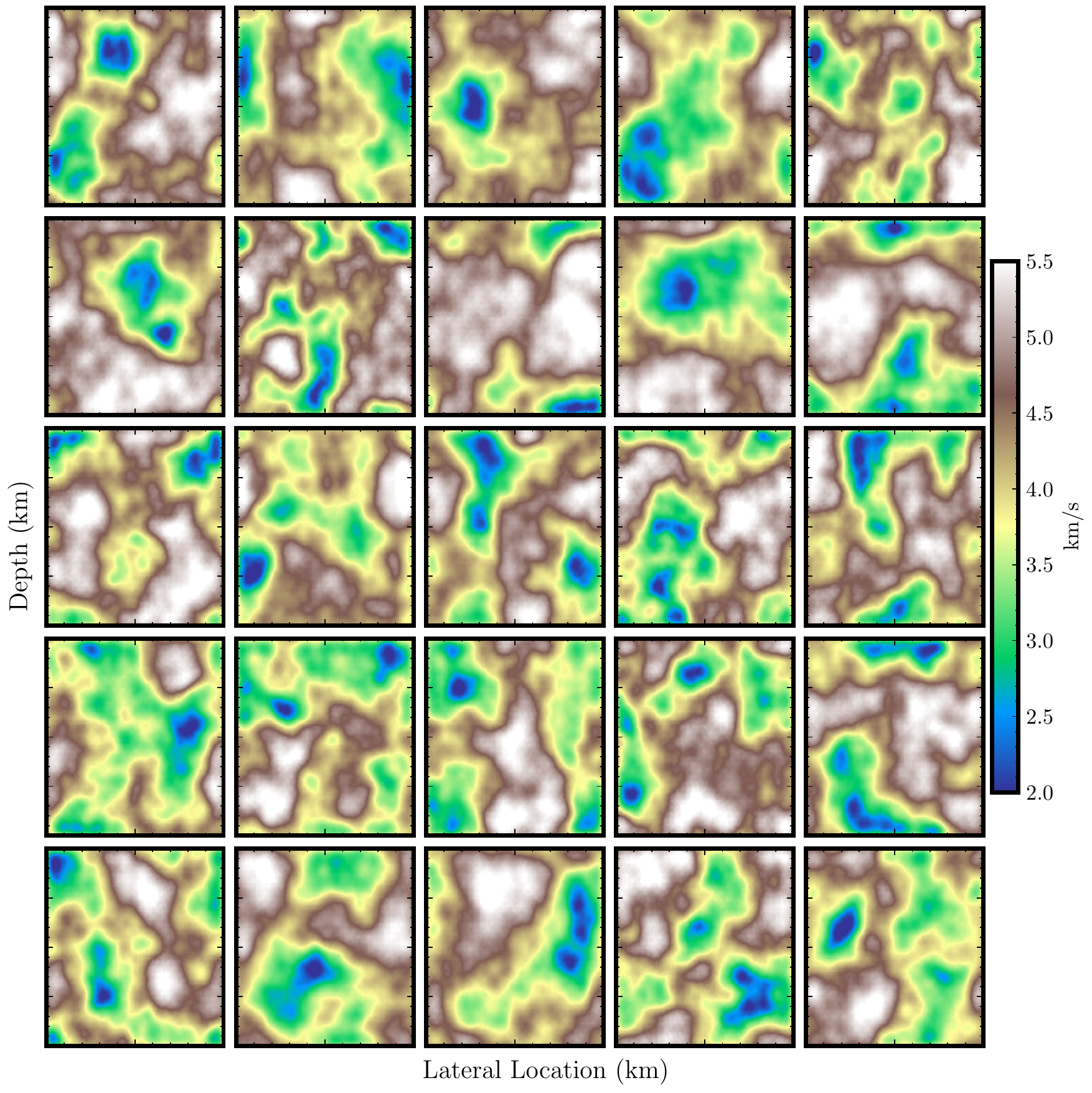}
    \caption{}
    \label{samplesa}
\end{subfigure}
\begin{subfigure}{.33\textwidth}
    \centering
    \includegraphics[width=0.939\linewidth]{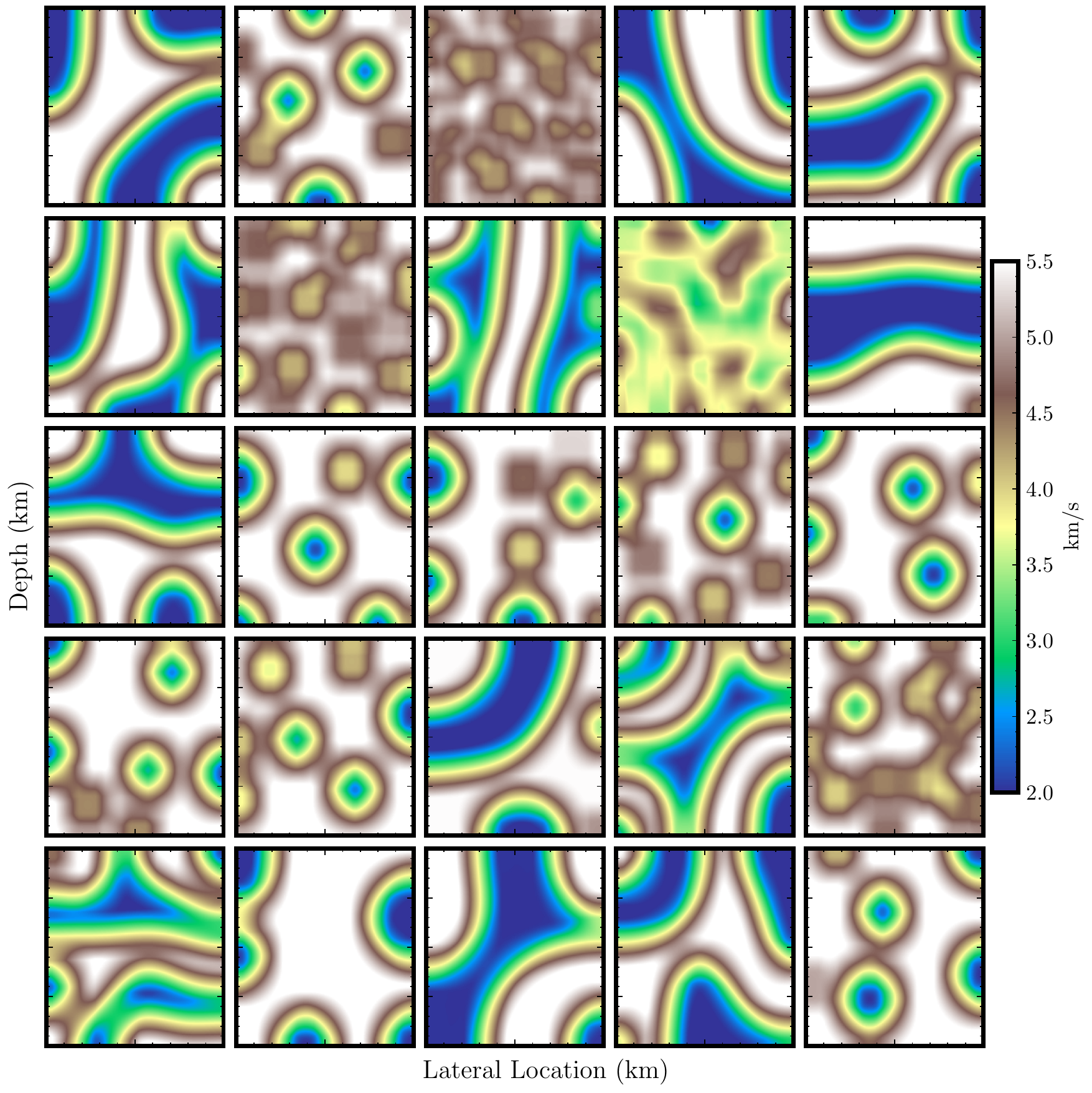}
    \caption{}
    \label{samplesb}
\end{subfigure}
\begin{subfigure}{.33\textwidth}
    \centering
    \includegraphics[width=0.939\linewidth]{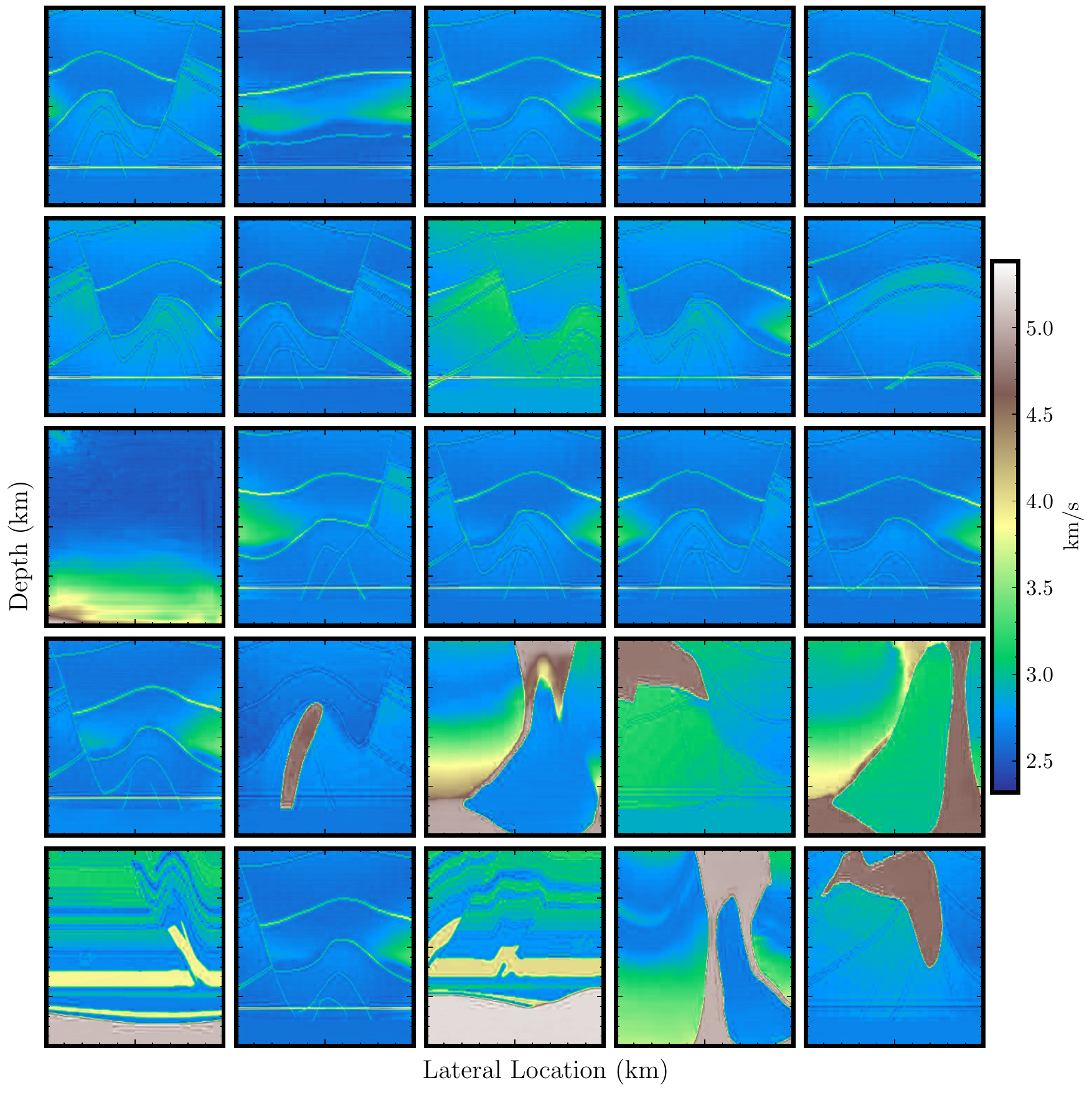}
    \caption{}
    \label{samplesc}
\end{subfigure}
\caption{
Samples from the latent diffusion model for the functional data representation task. Shown from left to right are the resulting analysis on different datasets; a) Gaussian random fields, b) Mechanical MNIST, and c) realistic Earth model.}
\label{samples}
\end{figure}

To perform this sampling, we extend the trained autoencoder with a diffusion model. Specifically, we utilize the recent latent diffusion model in which the diffusion (generative modeling) process is performed over the learned latent space. We utilize a diffusion model with 1000 diffusion steps, a linear noise scheduler with initial and final constants of 0.0015 and 0.0195, respectively, attention resolution of \{8,16\}, an initial learning rate of 2e-6, a channel multiplier of size \{1,2,3,4\}, a channel head of size 32, and a batch size of 5 image samples. The training process requires 27 epochs for the Gaussian spatial random fields, 32 epochs for the Mechanical MNIST dataset, and 289 epochs for the realistic Earth models. Figures \ref{samples} showcase the unconditional sample images from the trained latent diffusion model.

\section{Discussion}

Incorporating neural networks in solving PDEs is an emerging field in the scientific machine learning community. So far, two broad approaches have emerged, differing in whether the NN facilitates the learning process to replace either the PDE operator (neural operators) or a solution to a particular PDE (PINN). The latter offers more flexibility and can be done completely independently of conventional solvers. Thus, it offers features that motivate us to use it to replace the less flexible, mesh-oriented, conventional solvers. However, to realize this replacement, we need to improve the training efficiency of PINNs to at least rival the numerical ones. This is needed because, unlike neural operators, the training of PINNs is part of the inference, as PINNs are often trained for a fixed set of PDE parameters. Further solutions for new PDE parameters, unlike neural operators, require additional training. In other words, PINNs are trained \emph{deterministically} for a specific parametric PDE. To bridge this gap, orthogonal to previous approaches in speeding up the training process, latent representation learning is introduced by appending the conventional coordinates input with learned latent vectors representing certain PDE parameters. With this additional information, we encourage the neural network to use its interpolation capabilities (some of which shown in \cite{taufik2022upwind}) in its lower-dimensional manifold representation of the solutions to generate the corresponding solution. In other words, when such input latent vectors are sufficiently learned to compress and express the distribution of the PDE parameters, the trained PINNs will possess \emph{generative} features. By doing so, though we have to sacrifice an increase in the training cost compared to the usual PINN training, such an increase does not scale one-to-one with the number of representative PDE parameters needed to train our latentPINN. For example, as depicted in Figure \ref{ood}, the training cost for a hundred PDE parameters increases the training cost of the vanilla PINN (blue lines in Figure \ref{ood}) by 13 times at best (the GRF case) and 33 times (the mechanical MNIST) at worst. This increase is, however, justifiable, as once the latentPINN is trained, a new PDE solution can be predicted instantly for a sample PDE parameter from the trained distribution ($\sim$ 40k velocity fields) without additional NN training.

Our latentPINN implementation relies on the assumption of a well-represented PDE parameters distribution by its compressed latent representations. The main challenge in achieving such representations has been to harmonize the training samples reconstruction and generalization process to unseen samples. One significant limitation might arise when the new unseen PDE parameter lies far away from the training distribution. Our approach, like most of the neural operator models, will then either require re-training (from the first stage) or we resort to transfer learning to calculate the new PDE solution. In such scenarios, the calculated PDE solutions can be viewed as \emph{ansatz} for further PINNs training. With the recent development of latent diffusion models used in this study, we partially solve these trade-offs by promoting the use of an autoencoder to learn the compressed representation and diffusion models to sufficiently represent the posterior distribution of the PDE parameters. Specifically, a Kullback-Leibler divergence regularization is introduced for better generalization. Different means of regularization can also be used to achieve generalization during the latent compression \cite{esser2021taming}, which is worth investigating in future work.

As a byproduct of such a training scheme, not only does our framework facilitate PINN to learn a distribution of PDE parameters, its trained autoencoder can be further utilized for downstream tasks. One such task can be in the form of functional data representation. In this task, instead of facilitating surrogate maps for the PDE solution, the learned NN facilitates a functional representation for the infinite-dimensional vector spaces dubbed \emph{functional data}. Using the trained autoencoder as its backbone, a latent diffusion model can be deployed to perform generative modeling on such functional data. Thus, we obtain much more realistic drawn samples compared to the previous approach \cite[]{seidman2023variational}. Finally, in this study, we only consider unconditional sampling with such a model; further progress can include classes of the PDE parameters (i.e., geologic conditions for the Earth models or pattern classes for the mechanical MNIST dataset) to perform conditional sampling.

\section{Conclusion}

We proposed using learned compressed latent representations in physics-informed neural network training. More specifically, we utilize a latent diffusion model to learn compressed representations of the PDE parameters. These compressed representations are used as input samples, along with samples representing the coordinates of the solution space, to train a PINN to provide functional solutions of PDE for a range of PDE parameters. Though the proposed framework should comply with a more general class of parametric PDEs, we tested our algorithm on a first-order nonlinear PDE on three different datasets. With the additional latent vector input, the trained physics-informed neural networks are able to learn to generate PDE solutions as a function of PDE parameters. Thus, this approach retains the flexibility and accuracy features of the functional representation of PINN solutions while gaining the generalization abilities of a neural operator.

\section{Acknowledgement}

The authors thank King Abdullah University of Science and Technology (KAUST) for supporting this research and the Seismic Wave Analysis group for the supportive and encouraging environment. This work utilized the resources of the Supercomputing Laboratory at KAUST in Thuwal, Saudi Arabia.

\end{document}